\documentclass[journal,twoside,web]{ieeecolor}
\usepackage{tmi}
\usepackage{cite}
\usepackage{amsmath,amssymb,amsfonts}
\usepackage{algorithmic}
\usepackage{multirow}
\usepackage{booktabs}
\usepackage{textcomp}
\usepackage{makecell}
\usepackage{graphicx} %插入图片的宏包
\usepackage{float} %设置图片浮动位置的宏包
\usepackage{subfigure} %插入多图时用子图显示的宏包
\usepackage[colorlinks]{hyperref}
\usepackage[font=small]{caption}
\usepackage{dsfont}

\hypersetup
{
	pdfauthor = UnnamedOrange,
	colorlinks = true,
	linkcolor = black,
	anchorcolor = black,
	citecolor = black,
	urlcolor = black
}

\def\BibTeX{{\rm B\kern-.05em{\sc i\kern-.025em b}\kern-.08em
    T\kern-.1667em\lower.7ex\hbox{E}\kern-.125emX}}
\begin{document}
%\title{UniLearn: A Universal Vision Learner for 3D Medical Image Segmentation}
%\title{HybridMIM: A Hybrid MIM Framework for 3D Medical Image Segmentation}
\title{HybridMIM: A Hybrid Masked Image Modeling Framework for 3D Medical Image Segmentation}

\author{Zhaohu Xing, Lei Zhu,\IEEEmembership{Member, IEEE}, Lequan Yu, \IEEEmembership{Member, IEEE}, \\ Zhiheng Xing, and Liang Wan, \IEEEmembership{Member, IEEE}
\thanks{This paragraph of the first footnote will contain the date on which
you submitted your paper for review.  This work was supported in part by Tianjin Natural Science Foundation under Grant 21JCYBJC00510 and HKU Seed
Fund for Basic Research (Project No. 202111159073).} 
\thanks{Zhaohu Xing is with Medical College of Tianjin University, P. R. China (e-mail:xingzhaohu@tju.edu.cn). }
\thanks{Lei Zhu is with The Hong Kong University of Science and Technology, Hong Kong, and The Hong Kong University of Science and Technology (Guangzhou), China (e-mail: leizhu@ust.hk).}
\thanks{Lequan Yu is with The University of Hong Kong, Hong Kong (e-mail: lqyu@hku.hk).}
\thanks{Zhiheng Xing is with Haihe Hospital, Tianjin University, China (e-mail: 18920696025@189.cn).}
\thanks{Liang Wan is with Medical College, and the College of Computing and Intelligence, Tianjin University, China (e-mail:lwan@tju.edu.cn)}
}

\maketitle

\begin{abstract}
%Masked image modeling (MIM) with transformer backbones has recently been exploited, showing powerful potential as a self-supervised pre-training technique. 
Masked image modeling (MIM) with transformer backbones has recently been exploited as a powerful self-supervised pre-training technique. 
The existing MIM methods adopt the strategy to mask random patches of the image and reconstruct the missing pixels, which only considers semantic information at a lower level, and causes a long pre-training time.
This paper presents HybridMIM, a novel hybrid self-supervised learning method based on masked image modeling for 3D medical image segmentation.
Specifically, we design a two-level masking hierarchy to specify which and how patches in sub-volumes are masked, effectively providing the constraints of higher level semantic information. Then we learn the semantic information of medical images at three levels, including:
1) partial region prediction to reconstruct key contents of the 3D image, which largely reduces the pre-training time burden (pixel-level); 
2) patch-masking perception to learn the spatial relationship between the patches in each sub-volume (region-level).
and 3) drop-out-based contrastive learning between samples within a mini-batch, which further improves the generalization ability of the framework (sample-level). 
The proposed framework is versatile to support both CNN and transformer as encoder backbones, and also enables to pre-train decoders for image segmentation. 
We conduct comprehensive experiments on four widely-used public medical image segmentation datasets, including BraTS2020, BTCV, MSD Liver, and MSD Spleen. 
The experimental results show the clear superiority of HybridMIM against competing supervised methods, masked pre-training approaches, and other self-supervised methods, in terms of quantitative metrics, timing performance and qualitative observations. 
The codes of HybridMIM are available at 
\href{https://github.com/ge-xing/HybridMIM}{https://github.com/ge-xing/HybridMIM}. 
\end{abstract}


\begin{figure}[!t] %H为当前位置，!htb为忽略美学标准，htbp为浮动图形
\centering %图片居中
\includegraphics[width=0.5\textwidth]{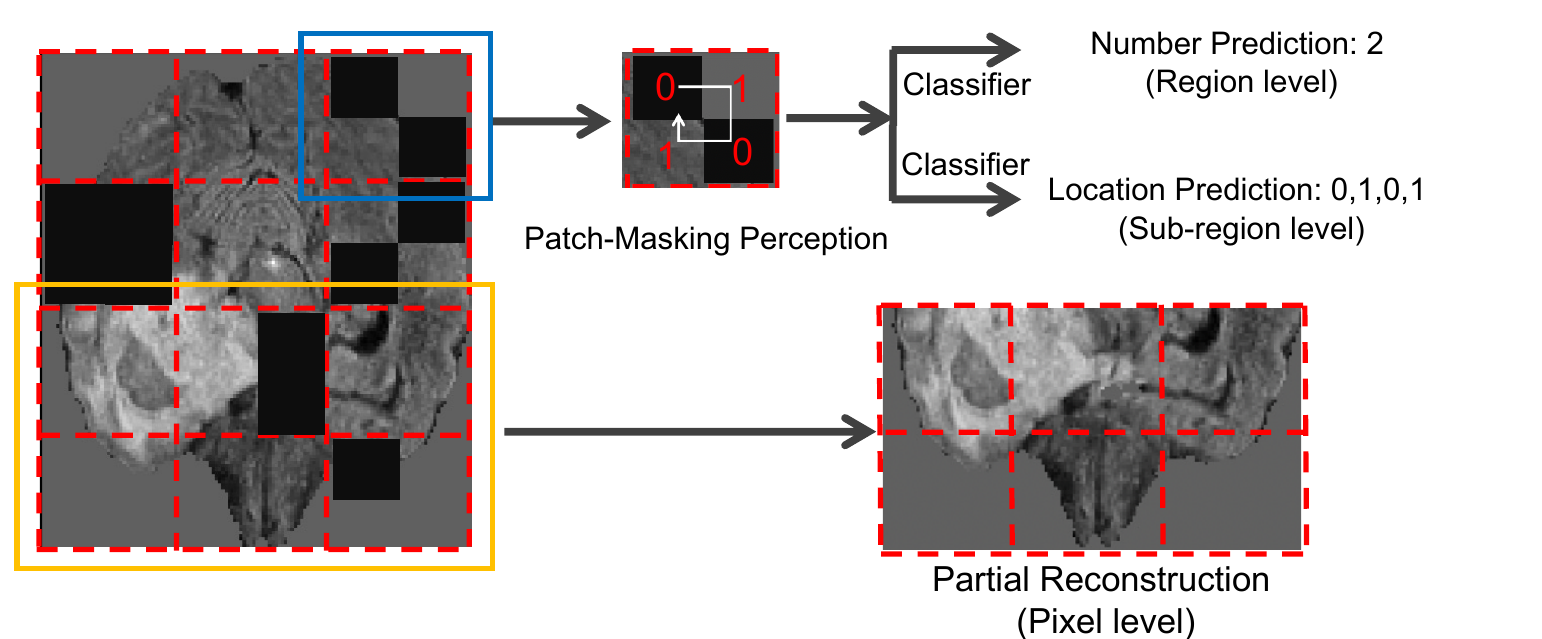} %插入图片，[]中设置图片大小，{}中是图片文件名
% 我们描述我们的想法以2D图像的形式。我们将整个图像分成多个子空间，在子空间中掩蔽部分更小的区域。我们以顺时针的方式对掩蔽与非掩蔽区域进行编码来得到子空间中掩蔽区域的数量与位置信息，并作为模型的预测目标。同时选择局部区域进行重构，有利于一个更快的预训练效率学习医学图像的空间解剖信息。此外，MP-SSL被设计不受网络架构的约束。
\caption{Illustration of our key idea in the 2D form. We regularly divide the input image into two-levels of patches, masking sub-regions or patches randomly. The masking information is encoded into binary, providing the locations and number of masked patches. In addition, local regions are selected for reconstruction, which facilitates a faster pre-training efficiency for high-dimensional medical images.}\label{fig:idea}
%We describe our ideas in the form of 2D images. We divide the whole image into several subspaces, masking partly smaller regions in the subspaces. We encode the masked and unmasked regions in a clockwise manner to obtain information on the number and location of masked regions in the subspace, which is used as a prediction target for the model. Also selecting local regions for reconstruction facilitates a faster pre-training efficiency to learn the spatial anatomical information of medical images. In addition, MP-SSL is designed to be unconstrained by the network architecture.} %最终文档中希望显示的图片标题
\label{Fig.main2} %用于文内引用的标签
\vspace{-3mm}
\end{figure}

\section{Introduction}
\label{sec:introduction}
% 在医学图像分析领域中，医学图像分割是一个热门且广泛应用的任务，它是非常有用的对于病灶的诊断与量化以及治疗的计划与评估。
\IEEEPARstart{I}n the field of medical image analysis\cite{litjens2017survey,khan2014survey,shamshad2201transformers}, automatic medical image segmentation remains a hot topic to facilitate lesion diagnosis and quantification, as well as treatment planning and evaluation.
% 近些年深度学习算法已经获得更为准确的分割结果，能更好的辅助专业医生进行判断，减轻医生审查MRI or CT影像的负担。
In recent decades, many deep learning-based algorithms for medical image segmentation have been developed to obtain improved segmentation performance. 
%, which can better assist medical professionals in making judgments and reduce the burden of reviewing magnetic resonance imaging (MRI) or computed tomography (CT) image.
%% 训练深度学习算法通常需要大量的人工标注数据，然而在医学图像分析领域，有标注的数据是十分稀缺和宝贵的。同时，可靠的标注需要多名专业医生协作，十分耗时且昂贵。
It is worth noting that training deep learning-based algorithms usually requires a large amount of manually labeled data. 
Yet in the field of medical image analysis, the amount of labeled 3D medical data is often small compared to natural images,
since labeling 3D medical data is not only tedious and time-consuming, but also needs intensive involvement of physicians to provide considerable domain expertise~\cite{tajbakhsh2020embracing}. 

%% 得不到充分的训练，有监督学习算法通常会达到性能瓶颈，无法更好的提升所处理任务的精度。
%Without sufficient training, supervised learning algorithms~\cite{zheng20153d,cai2021deep} usually reach performance bottlenecks and cannot better improve the accuracy of the processed tasks. 
% 为了克服有标注数据不充足的情况，自监督学习被提出利用无标注的语言/图像数据，使模型学习到领域内数据的通用表征。经过自监督学习后的模型在下游任务中finetune时拥有更快的收敛速度与更高的精度。
To alleviate the problem of insufficient labeled data, self-supervised learning (SSL) schemes~\cite{atito2021sit,dai2021up,liang2021swinir,doersch2017multi} 
have been developed to learn a generic representation of unlabeled data, and the obtained models can have faster convergence and higher accuracy when fine-tuned for downstream tasks.
%% 当前已经有一些有效的自监督学习方法被开发，例如自监督对比学习，掩蔽式语言/图像建模等。
% 其中对比学习通过构建正负样本对来学习不同样本之间的关系特征，而掩蔽式语言/图像建模则是通过掩蔽部分输入数据(语言/图像)，并让模型学习去重建被掩蔽的部分。
Typical self-supervised learning schemes include using tailored proxy tasks~\cite{gidaris2018unsupervised,zhou2021models}, self-supervised contrastive learning~\cite{gao2021simcse}, and masked image modeling~\cite{he2022masked}.
The former designs specific pretext tasks, such as inpainting, random rotation, distortion, for image restoration. 
Contrastive learning learns the relational characteristics between different samples by constructing positive and negative sample pairs. Masked image modeling, which is a recently developed technique, exhibits good potentials by learning the hidden context features through predicting what the masked parts of the input data should be. 
%reconstruct the masked parts by masking parts of the input data (language/image).
% 但是上述自监督学习方法专门为自然图像或自然语言开发。其与医学图像之间存在鸿沟，例如医学图像数据维度高，并且包含多种模态(computed tomography (CT) and magnetic resonance imaging (MRI))，导致无法很好的迁移到医学图像领域。
%the self-supervised learning methods described above were developed specifically for natural images or natural language. 

It is noted that most self-supervised learning methods are developed for natural images, and may not migrate well to the medical image domain~\cite{tang2022self}, since 3D medical image data has much higher dimensionality and can contain multiple modalities.
% 目前已经有一些方法探索如何将自监督学习方法应用于医学图像领域，例如有的方法利用聚类算法主动发现语义视觉词并将其作为自监督学习标签，使模型学习到了医学图像中丰富的语义信息，但是其预训练需要构建自监督学习标签，导致预训练步骤较多。
There have been several approaches to explore how to apply self-supervised learning to 3D medical images. For example, the visual word learning\cite{haghighi2021transferable} uses auto-encoder to actively discover semantic visual words and takes them as self-supervised learning labels, allowing the model to learn the rich semantic information. However, its pre-training requires the construction of self-supervised learning labels, resulting in more pre-training burden.
% 还有方法将掩蔽式图像建模与医学图像相结合，通过随机掩蔽输入图像的部分区域，来学习医学图像数据中的高维的空间解剖特征，但其模型结构具有局限性，而且使用全局重建导致预训练速度较慢。
UNetFormer~\cite{wang2022unetformer} utilizes a 3D Swin Transformer encoder, and directly applies masked image modeling by predicting randomly masked volumetric tokens. Just like current MAE~\cite{he2022masked} and SimMIM methods~\cite{xie2022simmim}, it recovers missing pixels, which are from a lower semantic level. What's more, reconstructing the whole image is time-consuming and leads to slow pre-training performance. 

%to learn high-dimensional spatial anatomical features in \textcolor{red}{medical image data by randomly masking parts of the input 3D image but its model structure has limitations. It does not test multiple model structures, and the use of global reconstruction leads to slower pre-training.}
% 在这个工作中，我们提出了一个混合的 MIM 自监督学习框架（Hybrid MIM）对3D医学图像分割问题，它通过多个特殊设计的子任务，从像素级别，区域级别，样本级别三个维度学习3D医学图像的语义表征。
%To alleviate the limitations of previous methods, 
In this paper, we propose a new hybrid self-supervised learning framework (HybridMIM) based on hierarchical masked image modeling for 3D medical image segmentation. 
Our framework learns the comprehensive semantic representation of 3D unlabeled medical images at pixel/region/sample three levels, 
%\ie, pixel level, region level and sample level.
%%
Specifically, as illustrated in Figure~\ref{fig:idea}, we design a two-level masking hierarchy to support masking sub-regions or patches randomly, and propose a joint self-supervised learning to account for semantic information at different levels.
At the pixel level, instead of predicting all the masked patches as MAE~\cite{he2022masked}, we utilize a partial region prediction scheme to reconstruct key parts of the 3D images, which can largely accelerate the pre-training.  
The masking status which reflects the locations and number of masked sub-regions, can help the model to characterize anatomical information at region-level. 
Finally, we utilize a dropout-based contrastive learning strategy at the sample level, to improve the model's capability of discrimination between different samples.

Our proposed HybridMIM is compatible with different networkd architectures, and we adopt UNet and SwinUNETR as the underlying architectures.
The effectiveness of our method is validated via two evaluation modes: pre-training and finetuning using the same data or using different data, on four downstream segmentation tasks, including BraTS2020~\cite{menze2014multimodal,bakas2017advancing}, BTCV~\cite{BTCV}, MSD-Spleen and MSD-Liver~\cite{antonelli2022medical}, which cover different organs, different object numbers and multi-modalities.
%BraTS2020 is a multi-modality brain MRI segmentation dataset, BTCV\cite{landman2015miccai} is an abdominal multi-organ segmentation dataset, MSD-Spleen is a spleen segmentation dataset, and MSD-Liver is a liver and liver tumor segmentation dataset.
%%
The comprehensive experimental results show that our HybridMIM outperforms existing state-of-the-art self-supervised methods in terms of both downstream segmentation task accuracy and training efficiency.

%We collected publicly available 1897 computed tomography (CT) images to pre-train a generic model (pre-training, finetuning using different data) that can perform transfer learning in different downstream tasks. In addition, we selected four publicly available datasets with different organs and modalities to pre-train task-specific models in each dataset (pre-training and finetuning using the same data). 

%% 为了验证我们提出的方法的通用性，我们选择了UNet与SwinUNETR架构作为backbone，利用HbridMIM方法预训练通用模型和不同数据集上的任务特定模型，实验结果表明我们提出的 HybridMIM 在训练效率，下游分割任务精度上比其他方法更有优势。
%To validate the enhancement of our proposed method for different architectures, we chose UNet and SwinUNETR as the underlying architectures and utilize the HybridMIM to pre-train generic and task-specific models on different datasets. The experimental results show that our proposed HybridMIM has advantages over other methods in terms of training efficiency and downstream segmentation task accuracy.
%% 此外，我们还进行了消融实验，证明了各个子任务的有效性。
%In addition, we also conduct ablation experiments to demonstrate the effectiveness of each subtask.
%% 总之，我们提出的基于掩蔽区域感知的自监督学习框架有以下几个特点：
%% 1. 协同的-实现了优秀的多任务协同学习能力通过子任务之间的促进。
%% 2. 通用的 - 适配不同的网络架构for更广泛的数据集。
%% 3. 高效的与鲁棒的 - 重建局部区域for 高效并且协同多任务for鲁棒.
In summary, our HybridMIM self-supervised learning method has the following dominant features:
\begin{itemize}
    \item Compatibility. Our method can serve as a general pre-training framework, and we demonstrate that it supports both multi-scale convolutional neural network and transformer network architectures.
    \item Comprehensiveness. Our method is designed to learn the semantic information within the data from pixel-level, region-level to sample-level.
    \item Robustness. Our method is proved to outperform SOTA methods in accuracy and training efficiency for multiple widely-used public segmentation datasets.
\end{itemize}
\section{RELATED WORK}

\begin{figure*}[htbp] %H为当前位置，!htb为忽略美学标准，htbp为浮动图形
\centering %图片居中
\vspace{-4mm}
\includegraphics[width=0.87\textwidth]{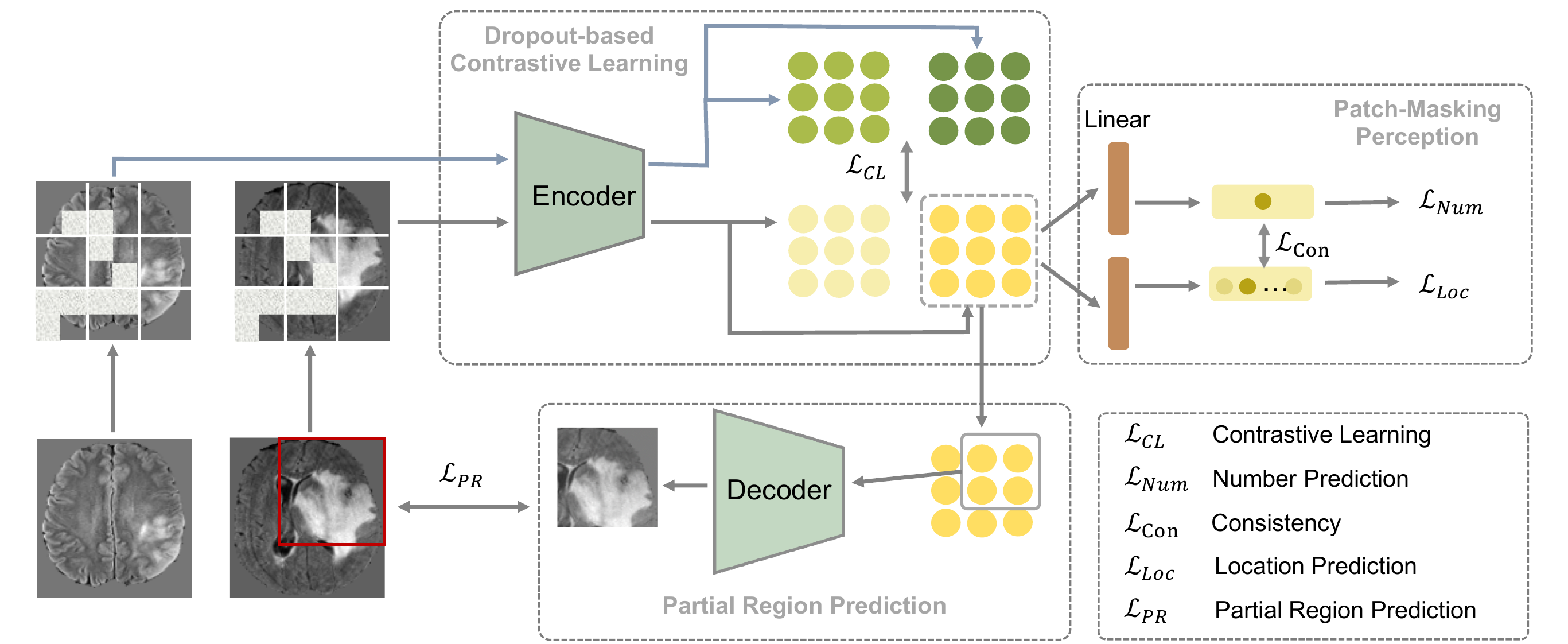} %插入图片，[]中设置图片大小，{}中是图片文件名
% 我们提出的masked-region perception多任务自我监督学习框架(MP-SSL)从全局与局部两个维度学习医学图像的通用表征。MP-SSL采用了一个两级的掩蔽方式，为了更好的对表征局部空间。以2D图像的形式来展示，被掩蔽后的图像经过两次编码，得到两组高水平特征。其中lcl代表对比学习损失，提升模型对不同样本特征全局的区分程度。lmr为掩蔽区域感知损失，共有三个损失函数，lmrnum，lmrpos，lmrconsis，分别代表掩蔽区域数量的损失，位置的损失和一致性损失。最后，通过解码手动选择的局部特征，重构对应的掩蔽区域。lrl表示重构损失，根据上下文信息重构空间中缺失区域，可以学习到丰富的医学图像解剖信息。
\caption{Overviwe of the HybridMIM pre-training framework. Input 3D medical images (demonstrated in 2D form) are randomly masked with a two-level masking strategy, then fed to the encoder twice to obtain two sets of feature representations. We use patch-masking perception, partial region reconstructions, and  dropout-based contrastive learning as proxy tasks to learn contextual representations of input images. } %最终文档中希望显示的图片标题
\label{Fig.main2} %用于文内引用的标签
\vspace{-3mm}
\end{figure*}

\subsection{SSL via Tailored Proxy Tasks}
%\subsection{Self-Supervised Learning via Tailored Proxy Tasks}
%% 该类方法通常通过设计pretext 任务来构造自监督学习标签。在自然图像处理领域中，有几个常用的pretext任务。图像修补学习视觉表征通过预测原始的图像patches。解决拼图游戏通过将图像patches随机打乱，并利用卷积神经网络进行还原，来学习图像的结构信息，此外，还有的方法通过将图像随机旋转，并使用网络预测旋转角度来增强对图像的理解能力。
This category of methods usually constructs self-supervised learning labels by designing pretext tasks. There are several commonly used pretext tasks in natural image processing. Image inpainting is adopted to learn visual representations by predicting the original image region~\cite{pathak2016context}. Jigsaw puzzles solving~\cite{noroozi2016unsupervised} learns structural information about images by randomly disrupting image patches and restoring them. 
In addition, random rotation estimation~\cite{gidaris2018unsupervised} enhances the understanding of images by randomly rotating images and using the network to predict the rotation angles.

% 在医学图像处理领域则包含更多样的pretext任务设计。
%% Models Genesis 利用多种方式扰乱原图像中部分像素值，例如distortion-based 方法打乱像素值 或 painting-based 方法屏蔽像素值。将被破坏的图像输入到encoder-decoder结构的网络进行还原，使网络学会重构初始正确的像素。这种设计可以使网络学习到医学图像数据的解剖特征。
One typical work using tailored proxy tasks in the medical imaging is Models Genesis\cite{zhou2021models}. 
%%
%such as distortion-based methods to disrupt pixel values or painting-based methods to mask pixel values. 
It uses a variety of ways, such as distortion or painting-based methods, to disrupt pixels in the original image\cite{gidaris2018unsupervised,pathak2016context,chen2019self}, and then utilize a network of encoder-decoder structure for restoration. 
This method  mainly focuses on the prediction of local information.
%, and does not consider global factors.
%% TransVW 定义医学图像中不同的区域并赋予不同的类别标签，称为语义视觉词。之后，TransVW随机将语义视觉词区域进行掩蔽，并通过encoder-decoder网络重建该区域并预测所属类别。
TransVW\cite{haghighi2021transferable} classifies different image regions as semantic visual words, and reconstructs randomly masked visual word regions while predicting the region's category.
The method achieves improved performance on downstream tasks, while it needs to find semantic visual words first, thus leading to more pre-training burden.
%% 然而，该方法需要首先寻找语义视觉词，因此导致预训练步骤较多。
%%
Yucheng Tang \textit{et al.} \cite{tang2022self} pre-trained swin transformers via three proxy tasks: inpainting, rotation and contrastive learning. 
Our method is built on the masked image modeling strategy, and constructs three proxy tasks at different levels, including predicting the partial masked region at pixel level, estimating the masking distribution at region level, and dropout-based contrastive learning at sample level.

\subsection{SSL via Contrastive Learning}
%\subsection{Self-Supervised Learning via Contrastive Learning}
%% 对比学习也是自监督学习中常用的方法，其核心在于利用不同的方式构建正样本与负样本，学习不同样本之间的关系特征。
%% 常见的构建正样本与负样本的方式为数据增强-将一个批次内相同样本经过不同的数据增强后作为正样本对，批次内不同样本作为负样本对。
%% 此外，SimCSE方法使用dropout构建正负样本。该方法首先将网络中的dropout层激活，然后将一个批量的数据前向计算两次，得到两组特征，因此来自相同样本的特征作为正样本，不同样本的特征作为负样本。该方法实现较好的性能提升且易于实现。
Contrastive learning is also a common approach in self-supervised learning, which centers on learning the relational features between samples by using different ways to construct positive and negative sample pairs.
A common way to construct sample pairs is data augmentation, in which one sample is augmented to form positive sample pairs, and different samples within a mini-batch are taken as negative sample pairs.
%%
%% 在医学图像分析领域中，Yucheng Tang\textit{et al.} 把基于数据增强的对比学习作为子任务之一，将来自相同子体积的增强样本作为正样本，不同子体积的样本作为负样本，并通过大量实验验证了其有效性。
In medical imaging, Yucheng Tang \textit{et al.} \cite{tang2022self} used contrast learning based on such data augmentation as one of the tailored proxy tasks, and had verified its effectiveness through numerous experiments.

% Krishna Chaitanya \textit{et al.}
%%
Additionally, in the NLP filed, the SimCSE method~\cite{gao2021simcse} proposes to use dropout to construct positive and negative samples. The method adds a dropout layer in the network and feed forward an input sample twice to obtain two sets of features. Features from the same sample are taken as positive samples, while features from different samples form negative samples. This scheme is easy to implement, and achieves good performance. 
We utilize SimCSE scheme to construct our constrative learning task.

\subsection{SSL via Masked Image Modeling}
%\subsection{Self-Supervised Learning via Masked Image Modeling}
%%% 此外，越来越多的工作正在探索基于屏蔽式图像建模的自监督学习任务使模型学习到全局的图像理解能力。例如Masked Autoencoders (MAE)在输入时丢弃掉被掩蔽的图像tokens，之后通过非对称的 transformer encoder-decoder 重构masked 区域，以此得到模型对于图像的多样化理解与表征。
More recently, a growing body of works is exploring to apply the idea of masked language modeling, which is successful in the NLP filed, to natural imaging. 
%%
%to enable models to learn global image understanding capabilities. 
%%
Masked Autoencoders (MAE) \cite{he2022masked}, built on ViT architecture, masks random image patches and reconstructs the missing pixels by an asymmetric transformer autoencoder.
To accelerate the pretraining speed, MAE discard masked tokens, which makes it less compatible with multi-scale processing expected in medical imaging. 
%%
%to obtain a diverse understanding and representation of the image by the model.
%% 同时，与MAE方法类似，SimMIM方法通过保留被掩蔽的图像tokens，使其可以应用于swin transformer，resnet等多尺度编码器。
Meanwhile, the SimMIM~\cite{xie2022simmim} achieves masked image modeling by utilizing multi-scale encoders such as Swin transformer and ResNet. SimMIM preserves the masked image tokens in the computation, and utilizes one convolutional layer as the decoder to increase timing performance, which may not handle well high-dimensional and high-resolution medical images.

% %% 由于MAE丢弃 masked tokens，采用非对称的 transformer encoder-decoder 加速提取图像特征，因此并不适用于通常处理医学图像数据的多尺度模型。
% Since MAE discards masked tokens and uses an asymmetric transformer encoder-decoder to accelerate the extraction of image features, it is not suitable for multi-scale models that usually deal with medical image data.
% %% SimMIM方法通过保留 masked tokens 的方式进行预训练，适配了SwinTransformer, ResNet等多尺度编码器。但是由于其编码的耗时长，因此选择简化解码器来做到性能权衡。与MAE方法使用8层Transformer 作为解码器相比，SimMIM使用了1层CNN结构作为解码器，但是简单的decoder并不能很好的处理高维和高分辨率的医学图像数据。
% The SimMIM\cite{xie2022simmim} method is pre-trained by retaining masked tokens and adapts multi-scale encoders such as SwinTransformer\cite{liu2021swin}, ResNet\cite{he2016deep}, etc. However, due to the long time consumption of its encoding, a simplified decoder is chosen to achieve the performance tradeoff. Compared with the MAE method that uses 8-layer Transformer as decoder, SimMIM uses 1-layer convolutional neural network as decoder, but the simple decoder is not able to handle high-dimensional and high-resolution medical image data well.

%%% 与在自然图像领域的自监督学习方法相比，UNetFormer 将掩蔽式图像建模的思想迁移到医学图像领域，随机掩蔽3D输入医学图像并使用encoder-decoder结构进行重建。它使用3D SwinTransformer结构作为编码器，多尺度的CNN或SwinTransformer结构作为解码器。然而，其结构是固定的，且在训练过程中需要极大的内存。
UNetFormer~\cite{wang2022unetformer} directly applies the idea of masked image modeling in medical imaging. It uses a 3D SwinTransformer structure as the encoder and a multi-scale CNN or SwinTransformer structure as the decoder, to reconstruct the randomly masked sub-volumes. However, its structure is fixed and requires an enormous amount of memory during the training process.
Our work extends the idea of masked image modeling to consider higher-level constraint on the semantic anatomical information. What's more, we accelerate the pre-training speed by using partial region prediction. 
\vspace{-1mm}
\section{Hybrid Masked Image Modeling}
The objective of model pre-training is to effectively encode anatomical information of the human body in unlabeled image data. 
%%
%Unlike previous works on masked image modeling~\cite{doersch2015unsupervised,chen2020generative,pathak2016context}, 
In this work, we propose a two-level masking strategy to facilitate the exploitation of semantic context within one image, from both pixel and regional views. 
To further improve the model's ability to discriminate between different samples, we apply dropout-based constrastive learning within a mini-batch.
Our pre-training is then accomplished with three proxy tasks, which learns the comprehensive semantic representation of images from pixel-level, region-level, and sample-level views.
Figure~\ref{Fig.main2} illustrates the idea, introduced next.

\vspace{-1mm}
\subsection{Masking Strategy and Network Architectures}
%\subsection{Two-level Masking Strategy and Network Architectures}
Assuming the input 3D medical image is $\mathcal{X} \in \mathcal{R}^{H\times W \times D}$, we first divide it equally into multiple non-overlapping sub-volumes (the first level), and then split each sub-volume along three dimensions equally, forming 8 smaller patches (the second level).
Let $R$ denote the number of first-level sub-volumes, and we will have $8R$ second-level patches in total.
Given the division, our masking strategy randomly selects second-level patches to do the masking (removing). 
Then, for each first-level sub-volume, we know how many and which second-level patches are masked. 
It is noted that the number and location information of the masked second-level patches can give a higher-level constraint on the anatomical information.
Hence, in addition to reconstruct the missing pixels, our HybridMIM also pays attention to the masking distribution so as to better learn the spatial anatomical information.

Following the auto-encoder architecture, our approach has an encoder that maps the input image data to the latent feature representation, and a decoder that reconstructs the original image from latent features. 
Since 3D medical images are multi-dimensional and have higher resolutions than natural images, we adopt two widely-used multi-scale segmentation models as the underlying infrastructure, including UNet~\cite{ronneberger2015u} and SwinUNETR~\cite{hatamizadeh2022swin}. 
For UNet that is a convolutional neural network, the masked image is feed into as a whole, while for SwinUNETR that is a transformer method, the masked image is split into tokens whose dimension is set the same as that of the second-level patches. 

\vspace{-1mm}
\subsection{Partial Region Prediction}
%%  SimMIM与UNetFormer方法通过解码器来重构掩蔽的区域学习图像的表征能力，SimMIM通过简化decoder加速自监督学习进程，但是在医学图像领域，由于医学图像数据是高维的和高分辨率的，通常需要一个多尺度的decoder重构图像。因此类似与UNetFormer，我们采用了多级的decoder进行重构，不同的是，我们只选择部分一级区域的high-level特征进行重构，而不是全局重构，大大提升了自监督学习的速度和对局部重要区域关注。
Previous masked image modeling methods~\cite{he2022masked,xie2022simmim} learn the feature representation of the image by reconstructing all the pixels values in the masked regions through a decoder.
SimMIM~\cite{xie2022simmim} accelerates the learning  by adopting a linear layer as the decoder. 
But considering that medical images are multi-dimensional and high-resolution, a multi-scale decoder is needed to reconstruct the image. 
To speed-up the pre-training process, we only select the features of some first-level sub-volumes (the target region) for reconstruction.
%% 局部重构分支被训练通过最小化重构区域与目标区域MASK像素的L2距离。
The partial reconstruction loss is defined as the $L_2$ distance between the reconstructed region and masked pixels in the target region,
\begin{equation}
\vspace{-2mm}
    \mathcal{L}_{\mathrm{PR}} = \frac{1}{|\hat{R}|}\sum_{r \in \hat{R}} || c_{r} - \hat{c}_{r}||_2,
\end{equation}
where $\hat{R}$ is a subset of sub-volumes in the target region, $|\hat{R}|$ is the number of involved sub-volumes, $c_{r}$ and $\hat{c}_{r}$ represent the prediction values and the input values, respectively.
Since the medical images usually have organs in the middle and may contain empty regions near the image boundary, we manually set the target region for one dataset around the image center. 

\subsection{Patch-Masking Perception}
Patch-Masking perception predicts the number and locations of masked patches for each sub-volume. 
This is achieved by adding two separate linear projection layers to the latent feature representation; see Figure~\ref{Fig.main2}.
As the masked patch number ranges within $[0,1,\ldots,8]$, we use a 9-dimensional softmax probability vector to represent the predicted number for $r$-th sub-volume, denoted as $u_r$.
Given the ground truth $\hat{u}$, a cross-entropy loss is used for number prediction task: 
\begin{equation}
    \mathcal{L}_{\mathrm{Num}} = - \frac{1}{R} \sum_{r=1}^{R} \hat{u}_{r} log( {u}_{r}) ,
\end{equation} 
% y代表模型对masked-region的数量的预测，g代表对应的ground truth。
where $R$ denotes the number of first-level sub-volume, $\hat{u}_r$ is represented as a one-hot vector. 

The predicted location $p_r$ is an 8-dimensional probability vector, with each element representing whether the corresponding patch is masked.
Here, we can employ the $\ell_0$ loss function\cite{han2022multimodal} for location prediction task, which counts non-zero elements in $p_r$, defined as:
\begin{equation}
\mathcal{L}_{\mathrm{Loc}}=\sum_{r=1}^R \sum_{k=1}^{8} s_r^k, \text { with } s_r^k=\left\{\begin{array}{ll}
1 & \text { if } \quad {p}_r^k \neq 0 \\
0 & \text { otherwise }
\end{array} ,\right.
\end{equation}
%在位置预测分支中，0表示第r个区域的第i个位置的预测，如果0=0，则此位置被预测为掩蔽区域，否则为非掩蔽区域，因此我们定义L0 损失函数：
where ${p}_r^k$ denotes the predicted probability of the $k$-th patch in the $r$-th sub-volume, and if ${p}_r^k$=0, this patch is predicted as a masked patch, otherwise it is a non-masked patch.
%% 考虑到l0 难优化，二元交叉熵损失函数被应用去估计一级区域中被掩蔽的二级区域的位置编码，公式如下，
Since $\ell_0$ loss is difficult to optimize, a binary cross-entropy loss function is utilized as an alternative:
\begin{equation}
\mathcal{L}_{\mathrm{Loc}} = - \frac{1}{R} \sum_{r=1}^{R} \sum_{k=1}^{8} \hat{p}_{r}^{k} log {p}_{r}^{k} + (1 - \hat{p}_{r}^{k}) log (1 - {p}_{r}^{k}),
\end{equation}
% I代表一级区域中二级区域的数量。
where $\hat{p}$ denotes the ground truth for the location coding of masked patches.

% 此外，在我们的方法中，对一级区域中掩蔽区域数量与位置的预测是各自独立的，但事实上，根据位置编码可以计算出对应的数量。因此，一致性损失被提出来避免位置编码计算得到的数量与预测的数量出现不一致的问题。
In addition, there is a straight-forward constraint between the predicted number and the predicted location information, in which the former can be calculated from the later. Therefore, we define a consistency loss to bridge them:
\begin{equation}
    \mathcal{L}_{\mathrm{Con}}=\frac{1}{2}(\sum_{r=1}^{R}{CE(u_r, \sum_{k=1}^{K}{\tilde{p}_r^k}) + MSE(\sum_{k=1}^{K}{p_r^k}, \tilde{u}_r))} ,
\end{equation}
where $CE$ and $MSE$ represent the cross entropy loss and mean squared error, $\tilde{p}_r^k = \mathds{1}[p_r^k > 0.5]$ gives an 0/1 estimate, and $\tilde{u}_r = argmax(u_r)$ gives the number estimate.

%% include the dataset list table
%\input{LaTeX/tables/dataset.tex}

\subsection{Dropout-based Constrastive Learning}

%% 在自然语言处理领域，SimCSE提出了将dropout作为最朴素的数据增强方式。将两个相同的句子前向计算两次，由于dropout的作用，会得到不同的句子表征。但此时两个句子表征来自于相同的句子，因此两个表征作为正样本。同时，SimCSE将一个batch内的其他句子表征作为负样本，SimCSE通过实验证明了这种方法可以简单有效的提升模型对于文本句子的表征能力。
%In the field of natural language processing, SimCSE\cite{gao2021simcse} proposes dropout as the most naive way of data augmentation. 
%% 类似的，为了提升模型对于3D医学图像数据的表征能力，我们将其思想应用于医学图像分析领域，将同一个图像前向计算两次得到的表征作为正样本，将批次内不同图像得到的表征作为负样本进行对比学习。
In order to further improve the model's generalization ability, we adopt dropout-based contrastive learning that is proven to be effective in the field of natural language processing\cite{gao2021simcse}.
Specifically, an masked image is encoded twice and two latent feature representations will be obtained due to the effect of dropout. 
Since they come from the same input, they are used as positive samples. 
The latent feature representations obtained from different images within a mini-batch are taken as negative samples for contrastive learning:
\begin{equation}
    \mathcal{L}_{\mathrm{CL}}=-\log \frac{\exp \left(\operatorname{sim}\left(v_i, v_j\right) / t\right)}{\sum_k^{2 N} \mathds{1}_{k \neq i, j} \exp \left(\operatorname{sim}\left(v_i, v_k\right) / t\right)},
\end{equation}% v表示正样本的表征之间的相似度。v表示负样本的表征之间的相似度。
where $t$ is the measurement of normalized temperature scale.
$\mathds{1}$ is the indicator function evaluating to 1 iff $k \neq i$. $v$ denotes the feature representation extracted by the encoder. $sim(v_i,v_j)$ denotes the similarity between the representations of positive samples, and  $sim(v_i,v_k)$ denotes the similarity between the representations of negative samples.

\vspace{-2mm}
\subsection{Loss Function}
%% HybridMIM将区域感知，局部重建，对比学习损失联合起来，进行多层次端到端的自监督学习。
Formally, we minimize a multi-objective loss functions combining masked region perception, partial reconstruction, and dropout-based contrastive learning losses, as follows:
\begin{equation}
    \mathcal{L}=\mathcal{L}_{\mathrm{PR}} + \lambda_{1}  \mathcal{L}_{\mathrm{Num}} + \lambda_{2} \mathcal{L}_{\mathrm{Loc}} + \lambda_{3} \mathcal{L}_{\mathrm{Con}} +  \lambda_{4} \mathcal{L}_{\mathrm{CL}},
\vspace{-2mm}
\end{equation}
where $\lambda_1,\lambda_2,\lambda_3,\lambda_4 $ are empirically set as 0.1, 0.1, 0.01, 0.1 respectively in all the experiments.

\section{EXPERIMENTS}
\subsection{Datasets}
% 为了充分的验证我们提出的MP-SSL方法的有效性，我们一共在4个3D医学图像分割数据集上进行实验，分别是BraTS2020，BTCV，MSD Liver，MSD Spleen。每个数据集的参数信息被展示在表1中。
%In order to fully validate the effectiveness of our proposed HybridMIM method, we conduct experiments on a total of four 3D medical image segmentation datasets, BraTS2020, BTCV, MSD Liver, and MSD Spleen, respectively. The parameter information of each dataset is presented in Table \ref{tab:data_intro}.

\textbf{Pre-training Dataset:}
We collected a total of 1897 CT images to construct our pre-training dataset. They come from 4 public CT image datasets, including the ATM22~\cite{atm22} (150 cases of chest), luna16~\cite{setio2017validation} (888 cases of lung), covid-19~\cite{roth2022rapid} (448 cases of lung) and FLARE21~\cite{MedIA-FLARE21} (411 cases of abdomen) datasets. 
%We split 20\% of each dataset for validation in the pre-training stage. 
% 80\% of the 1897 CT images collected for pre-training and 20\% for validation

%% BraTS2020数据集共包含369个病例的脑部图像，每个病例图像有4个模态分别是T1、T1Gd、T2、T2-FLAIR和3个分割目标分别为whole tumor (WT)、enhancing tumor (ET), and tumor core (TC)。由于输入图像尺寸较大，因此对每个病例我们会裁剪一块(128,128,128)的补丁在预训练和训练过程中。在预训练过程中，局部重构区域大小为（96，96，96）。
\textbf{BraTS2020 dataset}: BraTS2020 dataset~\cite{menze2014multimodal,bakas2017advancing} contains a total of 369 brain MRI images, and each case image has 4 modalities (namely T1, T1Gd, T2, T2-FLAIR) and 3 segmentation targets (WT: whole tumor, ET: enhancing tumor, TC: tumor core). All the data have been resampled to the same spacing (1.0, 1.0, 1.0). Due to the large size of the input image, we crop the training sub-volume of a size of (128,128,128). 

% BTCV数据集共包含30个病例的3D腹部多器官图像，每个病例图像有1个模态和13个器官分割目标。对每个病例我们会裁剪一块(96,96,96)的补丁在预训练和训练过程中。在预训练过程中，局部重构区域大小为（64，64，64）。
\textbf{BTCV dataset}: BTCV~\cite{BTCV} contains a total of 30 cases of 3D abdominal multi-organ CT images, with 1 modality and 13 organ segmentation targets per case. All cases are resampled to the same spacing (1.5, 1.5, 2.0). We crop the training sub-volumes of a size of (96,96,96).

% MSD Liver数据共包含201个病例的3D肺部图像，每个病例存在1个模态和2个分割目标(肺部和肺部肿瘤)，MSD Spleen数据集包含了61个病例的3D脾脏图像，每个病例有1个模态和1个分割目标(脾脏)。对每个病例我们会裁剪一块(96,96,96)的补丁在预训练和训练过程中。在预训练过程中，局部重构区域大小为（64，64，64）。
We further use two task datasets from MSD dataset~\cite{antonelli2022medical}. \textbf{MSD Liver} contains a total of 201 cases of 3D liver CT images, with 1 modality and 2 segmentation targets (liver, liver tumor) per case. The \textbf{MSD Spleen} contains 61 cases of 3D spleen CT images with 1 modality and 1 segmentation target (spleen) per case. For these two datasets, we resample all the cases to the same spacing (1.5, 1.5, 2.0) and crop the training sub-volumes of a size of (96,96,96).

\subsection{Evaluation Modes:} 
% 为了避免数据泄漏，预训练与训练数据保持一致。所有数据均切分80%作为预训练/训练集，20%作为验证集。
To make a comprehensive comparison, we adopt two evaluation modes to do pre-training and finetuning:
%on the above four downstream segmentation tasks.
%, including BraTS2020, BTCV, MSD-Spleen and MSD-Liver.
\begin{itemize}
    \item Generic pre-training mode: A generic model is pre-trained on our collected pre-training dataset, and finetuned on the four segmentation tasks.
    \item Task-specific pre-training mode: A task-specific model is pre-trained and finetuned on each segmentation task dataset. To avoid data leakage, we do self-supervised pre-training on the training set of the downstream task datasets, which are all split 80\% for training.
\end{itemize}
In finetuning stage for segmentation, we utilize the pre-trained encoder and decoder by replacing the last layer with a new random $c \times 1 \times 1 \times 1$ convolution layer, where $c$ is the number of segmentation targets.

\subsection{Implementation Details}
%\textbf{Basic setup for pre-training.}
% 我们通过多组对照实验选出 最优的一级区域，二级区域与重建的局部区域的大小。看 V. Result, section A for details. 
The default options for the components of HybridMIM are: the two-level masking strategy with a mask ratio of 0.4, the first-level sub-volume size of $32\times 32 \times 32$, the second-level patch size of $16\times 16 \times 16$; the partial region prediction with the reconstructed region size of $96\times 96 \times 96$ for BraTS2020, and $64\times 64 \times 64$ for BTCV, MSD Liver and MSD Spleen. 
These settings are determined via multiple control experiment; See Figure~\ref{fig:pretraining_setting} for details. 
%%
% 对于 MSD 肝脏、MSD 脾脏和 BTCV 实验，我们使用分辨率为 96 × 96 × 96 的随机裁剪图像，并且预训练实验使用每个 GPU 4 个批量大小（使用 $96\times96\times96$ 补丁和 $64\times64 \times64$ 解码大小）。对于 BraTS2020 数据集，我们使用128作为随机裁剪大小，并且每个GPU使用2个批量大小。
For the pre-training dataset, BTCV, MSD liver and MSD spleen, we use randomly cropped images with a resolution of 96 × 96 × 96 and a batch size of 4 per GPU. For BraTS2020 dataset, we use $128\times128\times128$ as the random crop size and a batch size of 2 per GPU. As each image case in BraTS2020 contains 4 modalities, we concatenate each modality in channel dimension at the input of the network.

% 我们使用Pytorch1.12.1-cuda11.3与Monai1.0.0作为基本框架。使用4 个NVIDIA A100 Tensor Core GPU和2个NVIDIA Tesla V100 GPU作为运行环境。
Our model is implemented in Pytorch 1.12.1-cuda11.3 and Monai 1.0.0. 
%所有实验均应用随机翻转、旋转、强度缩放和移位的数据增强变换。优化器使用AdamW以及 1e-4 的初始学习率、 1e-5 的衰减和50个epochs的warmup。
In both pre-training and finetuning, we use an AdamW\cite{loshchilov2017decoupled} optimizer along with a $cosine$ learning rate scheduler (an initial learning rate of 1e-4, a decay of 1e-5, and a warmup\cite{he2016deep} of 50 epochs).
% 对于BraTS2020数据集，我们共运行300个epoch，对于BTCV数据集，我们共运行2000个epoch，对于MSD Liver与Spleen数据集，运行600个epoch。
We run a totoal of 10w step for all pre-training experiments, and
in finetuning, we run 300 epochs for the BraTS2020 dataset, 2000 epochs for the BTCV dataset, and 600 epochs for the MSD Liver and Spleen datasets.
No data augmentation is applied at the pre-training stage, while a light strategy is used in the finetuning: random flip, rotation, intensity scaling and shifts with probabilities of 0.2, 0.2, 0.1, and 0.1, respectively.
%with a probability of 0.2 in each dimension, random rotations 90 degree with a probability of 0.2, intensity scaling and shifts with a probability of 0.1.  
%%
All experiments are conducted on a cloud computing platform with four NVIDIA A100 Tensor Core GPUs and two NVIDIA Tesla V100 GPUs.

% 此外，由于对比学习需要正负样本，批量大小大于1并且网络中必须添加dropout层。
%In addition, since comparison learning requires positive and negative samples, the batch size is larger than 1 and a dropout layer must be added to the network.

\begin{table*}[t]
    \vspace{-3mm}
    \caption{Quantitative comparison on BTCV muti-organ sementation dataset. Note: Spl: spleen, RKid: right kidney, LKid: left kidney, Gall: gallbladder, Eso: esophagus,Liv: liver, Sto: stomach, Aor: aorta, IVC: inferior vena cava, PSV: portal and splenic veins, Pan: pancreas, Rag: right adrenal glands, Lag: left adrenal glands. The task-specific pre-trained models are marked with *.}
    \centering 
    \label{tab:btcv_segmentation} %\footnotesize
    \renewcommand\arraystretch{1.3}
    \setlength\tabcolsep{4pt}%调列距
    \resizebox{0.9\textwidth}{!}{
    \begin{tabular}{c | c | c c c c c c c c c c c c c }
    % \toprule[1pt]
    % Dataset & \multicolumn{11}{c}{BraTS2020} \\
    % \hline
    % \multirow{2}{*}{Methods} & \multirow{2}{*}{\makecell{Param\\(M)}} & \multirow{2}{*}{\makecell{FLOPs\\(G)}} &  & \multicolumn{2}{c}{WT} &  & \multicolumn{2}{c}{TC} & &  \multicolumn{2}{c}{ET} &  & & \multicolumn{2}{c}{Ave} \\
    % \cline{5-6} \cline{8-9} \cline{11-12} \cline{15-16} 
    \hline
    Methods & Avg & Spl & RKid & LKid & Gall & Eso & Liv & Sto & Aor & IVC & PSV & Pan & Rag & Lag \\
    \hline
    Segresnet & 81.29 & 94.55 & 93.35 & 93.41 & {\color{green}75.59} & 73.44 & 95.96 & 80.89 & 89.00 & 84.24 & 71.48 & 79.12 & 65.51 & 60.07 \\
    
    UNETR & 81.33 & 94.66 & {94.27} & 94.09 & 65.23 & 74.20 & {\color{red}96.90} & 80.06 & 89.16 & 84.04 & {73.46} & 80.32 & 64.30 & {\color{green}66.65} \\
    SwinUNETR &81.81 & 94.72 & 94.23 & 93.89 & 66.60 & 74.54 & 96.63 & 78.77 & 89.79 & 83.64 & {\color{blue}74.69} & {\color{red}82.19} & {67.76} & {66.18} \\
    \hline
     ModelGen & 81.45 & 91.99 & 93.52 & 91.81 & 65.11 & {\color{red}76.14} & 95.98 & {\color{red}86.88} & 89.29 & 83.59 & 71.79 & {81.62} & {\color{green}67.97} & 63.18 \\
    TransVW & {82.27} & {95.56} & 94.20 & {\color{red}94.59} & 70.42 & 73.25 & 96.51 & {\color{blue}85.65} & {\color{blue}90.44} & {\color{blue}85.80} & 73.19 & {\color{blue}81.91} & 66.17 & 61.62\\
    UNetFormer* & 82.18 & 94.27 & 94.15 & 93.80 & {\color{blue}75.86} & {75.05} & {96.72} & 81.74 & 90.13 & 83.32 & 72.41 & 79.90 & 67.06 & 63.95 \\
    UNetFormer & 82.44 & {\color{green}95.90} & {\color{red}94.61} & 94.28 & 71.51 & 75.08 & 96.51 & 81.46 & 90.06 & 85.84 & {\color{red}75.34} & 80.72 & {\color{blue}68.50} & 61.78\\
    
    \hline
    HybridMIM*(Swin) & {82.41} & {\color{red}95.95} & {\color{blue}94.60} & {\color{green}94.36} & 65.79 & {\color{green}75.46}  & {\color{green}96.74} & 82.54 & {90.14} & {84.80} & {\color{green}74.02} & 80.36 & {67.87} & {\color{red}68.47}\\
    HybridMIM*(UNet) & {\color{green}82.62} & {95.27} & {94.25} & {94.15} & {\color{red}78.67} & 74.24 & {96.68} & {\color{green}83.31} & {\color{green}90.25} & {\color{red}85.82} & 73.07 & 80.24 & 65.02 & 62.94 \\
    \hline
    HybridMIM(Swin) & {\color{blue}82.63} & {\color{blue}95.95} & 94.25 & 94.26 & 72.56 & 74.14 & {\color{blue}96.78} & 79.25 & 90.17 & 85.10 & 73.64 & {\color{green}81.83} & {\color{red}69.21} & {\color{blue}66.88}\\
    HybridMIM(UNet) & {\color{red}83.00} & 95.68 & {\color{green}94.43} & {\color{blue}94.40} & 74.33 & {\color{blue}75.84} & 96.72 & 82.92 & {\color{red}90.86} & {\color{red}86.43} & 72.96 & 81.16 & 66.98 & 66.25\\
    \hline
    \end{tabular}
    }
\end{table*}

% \subsection{Evaluation Metrics}
% $Dice$ score and 95\% Hausdorff Distance ($HD 95$) are adopted for quantitative comparison. $HD 95$ is based on the calculation of the $95^{th}$ percentile of the distances between boundary points in $X$ and $Y$. 
% \begin{equation}
% Dice=\frac{2|A \cap B|}{|A|+|B|},
% \end{equation}

% \begin{equation}
% HD=\max \left\{\sup _{x \in X} \inf_{y \in Y} d(x, y), \sup _{y \in Y} \inf_{x \in X} d(y, x)\right\},
% \end{equation}

% where A and B denote the ground truth and prediction of
% voxel values. X and Y denote ground truth and prediction surface point sets. $\sup$ represents the supremum, $\inf$ the infimum.

\subsection{Benchmarking}

% 为了使实验结果更有说服力，MP-SSL共与其他六种方法进行对比。这些方法中包含了不同的网络架构，同时也包含了SOTA的监督学习方法与自监督学习方法。
For a thorough evaluation, HybridMIM is compared with SOTA SSL methods as well as fully supervised learning methods, which both cover  CNN and transformer architectures. 

\textbf{Self-supervised methods:} We compare HybridMIM with Models Genesis~\cite{zhou2021models}and TransVW~\cite{haghighi2021transferable}, which are the most recent multi-task SSL methods for 3D medical imaging. 
We also examine an masked image modeling-based SSL method, UNetFormer~\cite{wang2022unetformer}, which is built upon the SwinTransformer architecture. 
% 对于TransVW和ModelGenesis方法，我们使用的是官方开源的权重。对于UNetFormer方法，我们复现了两种预训练模型，UNetFormer*表示在我们收集的1897个CT图像上预训练的通用模型，而UNetFormer表示任务特定的预训练模型，预训练数据与下游任务一致。
As ModelGenesis and TransVW officially release their pre-trained model weights, we skip the pre-training step and conduct finetuning on the four downstream segmentation tasks. 
%%
%Note that since these two methods both utilize a pre-training dataset of 5050 CT images, larger than ours, it is reasonable to start with their models for finetuning.
%%
For UNetFormer, we use its public codes and experiment with it in both evaluation modes. 

%replicate two pre-trained models. UNetFormer* denotes the generic model pre-trained on our collection of 1897 CT images, while UNetFormer denotes the task-specific pre-trained model with pre-trained data consistent with the downstream task.

%%
% 对于监督学习方法，我们选择了SegresNet，UNETR和SwinUNETR作为对比方法。其中SegresNet是一个基于卷积神经网络且性能良好的架构。UNETR与SwinUNETR分别采用ViT transformer与SwinTransformer结构作为编码器，这可以更好的建模全局特征。他们均是最近表现良好的3D医学图像分割方法。
\textbf{Supervised methods:}
We also make comparison with state-of-the-art supervised segmentation methods in medical imaging. 
SegresNet~\cite{myronenko20183d} is a CNN-based architecture with good performance. 
UNETR~\cite{hatamizadeh2022unetr} and SwinUNETR~\cite{hatamizadeh2022swin} are the most recent transformer-based methods for 3D medical image segmentation, which use vision transformer and SwinTransformer structures as encoders, respectively. Here, we use their public codes in the experiments.
%%
% 此外，HybridMIM适用于不同的网络架构。因此我们分别使用UNet和SwinUNETR作为基础架构，并且分别预训练通用模型与任务特定模型，来验证HybridMIM方法对其性能的提升。
%Moreover, HybridMIM applies to different network architectures. Therefore, we use UNet and SwinUNETR as the infrastructure and pre-train the generic and task-specific models, respectively, to verify the performance improvement of the HybridMIM approach.

\section{Results}

% 这个部分展示了我们结果的性能优势，首先，我们通过多组对照试验确定了最优的架构参数，然后我们基于UNet与SwinUNETR预训练了通用模型与任务特定模型，在结果表中，*表示通用模型，否则为任务特定模型。同时，我们使用了多种来源的数据，涉及了不同模态，不同器官和不同的分割目标来验证HybridMIM的鲁棒性。此外，我们还验证了不同有标签数据比例下，HybridMIM依然能够有较高的性能优势。最后，我们还进行了消融实验，验证了HybridMIM中不同模块的有效性。
This section demonstrates the significance of our proposed HybridMIM method. 
First, we make comparison with the current state-of-the-art approaches from four aspects: downstream segmentation performance (quantitative and qualitative), annotation cost reduction, and pre-training speed. 
We then conduct ablation experiments to explain how to determine the optimal architectural parameters, and illustrate the contribution of each component to the performance of HybridMIM.

%This section demonstrates the performance advantage of our results. First, we determine the optimal architectural parameters by multiple controlled trials, and then we pre-train the generic and task-specific models based on UNet with SwinUNETR. In the result table, * indicates the generic model, otherwise the task-specific model. Also, we use data from multiple sources involving different modalities, organs, and segmentation targets to validate the robustness of HybridMIM. In addition, we verify that HybridMIM can still have high-performance advantages with different scales of labeled data. Finally, we also conduct ablation experiments to validate the effectiveness of different modules in HybridMIM.

\begin{table}[th]
    %\centering
    % 其中MSD Liver数据集需要分割肝脏和对应的肿瘤。MSD Spleen数据集需要分割脾脏。我们使用Dice和HD95来评估不同对比方法的性能。无论基于UNet架构还是SwinTransformer架构，MP-SSL方法都对其有很高的性能提升，并实现了state-of-the-art的结果。
    %\vspace{-2mm}
    \caption{The MSD Liver dataset requires segmentation of the liver and the corresponding tumor. and the MSD Spleen dataset requires segmentation of the spleen.}
    \label{tab:msd_segmentation}
    \renewcommand\arraystretch{1.3}
    \setlength\tabcolsep{3pt}%调列距
    \resizebox{\columnwidth}{!}{
    \begin{tabular}{c | c c c c c c | c c c}

    \hline
    Organ & \multicolumn{6}{c}{Liver} & \multicolumn{2}{c}{Spleen} \\
    \hline
    Metrics & Dice & Dice & Dice & HD & HD & HD & Dice & HD \\
     & liver & tumor & Avg & liver & tumor & Avg &  &  \\
    \hline
    SegresNet & 95.53 & 48.26 & 71.90 & 0.81 & {15.31} & 25.31 & 94.10 & 0.5\\
    UNETR & 93.07 & 33.59 & 63.33& 1.26 & 30.50 & 15.88 & 94.04 & 0.58\\
    SwinUNETR & 95.14 & 45.11 & 70.13 & 0.89 & 21.31 & 11.11 & 94.61 & 0.25\\
    \hline
    ModelGen & 95.22 & {52.53} & 73.87 & 0.67 & 18.83 & 9.75 & 94.43 & 0.63 \\
    TransVW & 95.67 & 52.10 & 73.88 & 0.60 & 21.36 & 10.98 & 95.55 & 0.41 \\
    UNetFormer* & 95.50 & 49.81 & 72.65 & {0.52} & 21.72 & 11.12 & 95.36 & 0.25 \\
    UNetFormer & 95.83 & 50.25 & 73.04 & 0.43 & 18.66 & 9.55 & 95.59 & 0.30 \\
    
    \hline
    HybridMIM*(Swin) & 95.45 & 50.19 & 72.82 & 0.69 & \textbf{15.21} & \textbf{7.95} & 95.87 & 0.25\\
    HybridMIM*(UNet) & \textbf{96.35} & 52.38 & \textbf{74.36} & 0.59 & 19.98 & 10.28 & {95.94} & \textbf{0.20} \\
    \hline
    HybridMIM(Swin) & 95.86 & 50.45 & 73.16 & 0.42 & 17.36 & 8.89 & 95.97 & 0.20 \\
    HybridMIM(UNet) & 95.70 & \textbf{52.81} & 74.26 & \textbf{0.27} & 18.25 & 9.26 & \textbf{96.05} & \textbf{0.20} \\
    \hline 
    \end{tabular}
    }
    \vspace{-2mm}
\end{table}

\subsection{Quantitative Comparison to Previous Methods} 
\textbf{BTCV multi-organ segmentation.} The multi-organ segmentation results are listed in Table \ref{tab:btcv_segmentation}, in which
the first, second, and third best dice scores are marked in red, blue, and green colors, respectively. 
Among the comparative methods, we can see that those with self-supervised pre-training generally achieve averagely better results than those fully supervised methods. 
TransVW obtains the best average Dice of 82.27\%,  
while for UNetFormer, its generic pre-trained model presents an average Dice of 82.44\%, outperforming the task-specific pre-trained model UNetFormer* by 0.26\%. 

% 与其他对比方法相比，我们的基于UNet和SwinTransformer架构的方法均取得了有竞争力的结果。红色，蓝色，绿色分别代表最高的dice得分，第二高的dice得分与第三高的dice得分。可以清楚的发现，基于SwinUNETR架构的任务特定模型Swin(HybridMIM)在7项指标中均位于前三名，实现了82.41%的Dice平均值。而基于UNet架构的通用预训练模型UNet*(HybridMIM)，在4项指标中位于前两名，相比于其他方法实现了最高的平均Dice，83.00。在BTCV多器官分割任务中，通用预训练模型的性能均高于任务特定预训练模型。
In comparison, our methods on both UNet and SwinTransformer architectures outperform most SOTA methods, and the generic pre-trained models get better performance than their task-specific pre-trained counterparts.  
Specifically, the generic pre-trained model HybridMIM(UNet) presents the highest average Dice of 83.00\%,
%We can find that the task-specific model Swin (HybridMIM) based on SwinUNETR architecture is in the top three in all seven metrics, achieving an 82.41\% Dice average. 
% 拿性能最好的UNet*(HybridMIM)来说，它实现了最高的83.0%的平均Dice，比表现较好的同样在通用数据集上预训练的UNetFormer*模型提升了0.56%。并且UNet*(HybridMIM)在13个分割目标中有9个目标的分割结果均优于UNetFormer*。
which is 0.56\% better than the best SOTA model UNetFormer, and outperforms it in 9 out of 13 segmentation targets.
% 并且基于SwinUNETR架构的任务特定预训练模型在Lag器官上分割效果明显优于其他对比方法，达到了68.47%的dice值，比第二名UNETR高出1.82%。而基于UNet架构的任务特定预训练模型在Gall器官上分割效果显著，达到了 the dice of 78.67%，而第二名UNetFormer与第三名Segresnet方法的dice均没有超过76%。
%
Furthermore, the task-specific pre-trained model HybridMIM*(Swin) segmented significantly better than the other methods on the Lag organ, reaching the Dice of 68.47\%, which is 1.82\% higher than the second place UNETR, while HybridMIM*(UNet) reports a significantly better result on the Gall organ, reaching a Dice of 78.67\%. 
%In comparison, neither the second-place UNetFormer nor the third-place Segresnet method had more than 76\% Dice.

% 肝脏与肝脏肿瘤分割结果被展示在表3的左侧。加粗字体表示最优的指标。可以清晰的看到，我们提出的基于UNet架构的任务特定预训练模型UNet(HybridMIM)在肝脏的分割上有最好的Dice of 96.35%，比第二名TransVW提升了0.68%。同时其在肝脏肿瘤的分割中达到了Dice of 52.38%，仅次于ModelGen方法的52.53%。此外，UNet(HybridMIM)也实现了两个分割指标的最好的平均Dice，为74.36，比第二名TransVW方法提升了0.48%。
\textbf{Liver and liver tmuor segmentation.} As shown in Table \ref{tab:msd_segmentation}, 
%The bolded font indicates the best metrics.
our task-specific pre-trained model HybridMIM*(UNet) achieves the best average Dice of 74.36\%, with an improvement of 0.48\% over the second-place TransVW method.
Furthermore, it reports the best Dice of 96.35\% for the segmentation of the liver, which is 0.68\% better than the second place TransVW; and obtains a Dice of 52.38\% in the segmentation of liver tumors, only slightly lower than the second place ModelGen method with 52.53\%. 
% 对于HD95分割指标，基于UNet(HybridMIM)在肝脏的分割中位于第二名，HD95结果为0.59，略高于UNetFormer方法的0.52。在肝脏肿瘤的分割中为第三名，HD95为19.98。
For the HD95 segmentation metric, the HybridMIM*(UNet) gets an average HD95 of 10.28, ranked in the third place.
%is in second place in the segmentation of the liver with an HD95 result of 0.59, slightly higher than the UNetFormer method of 0.52. It was in third place in the segmentation of liver tumors with an HD95 result of 19.98, and the average HD95 was also in third place.
% 同时，Swin(HybridMIM)总体来说在HD95指标上表现更好。其在肝脏肿瘤的分割上拥有最好的HD95，为15.21，并且其在肝脏与肝脏肿瘤两个分割目标上实现了最好的的平均HD95，为7.95，比第二名ModelGen方法降低了1.8。相比于没有经过预训练SwinUNETR方法，Swin(MP-SSL)有更加明显的提升。其在肝脏与肝脏肿瘤的平均Dice得分达到了72.82%，比SwinUNETR方法提升了2.17%。
%Meanwhile, the Swin(HybridMIM) performed better overall on HD95 metrics. 
%It achieves the best HD95 of 15.21 for liver tumor segmentation and the best average HD95 of 7.95 for liver and liver tumor segmentation targets, which is 1.8 lower than the ModelGen method in second place. 
In addition, compared to the SwinUNETR method without pre-training, both HybridMIM*(Swin) and HybridMIM(Swin) which employ SwinUNETR as the underlying architecture, have more significant improvements in all the metrics. 
%%
%HybridMIM*(Swin) and HybridMIM(Swin) get an average Dice score of 72.82\% and 73.16\%, 2.69\% and 3.03\% higher than the SwinUNETR method, respectively.

% 脾脏的分割结果被展示在表3的右侧。可以看到，基于UNet与SwinUNETR架构的HybridMIM均表现出了优秀的性能，无论是在Dice还是在HD95上。基于UNet*(HybridMIM)获得了 state-of-the-art 的Dice与HD95，分别为96.05与0.20，在Dice得分上相比于同样表现较好的对比方法TransVW提升了0.50%，比基于Transformer架构的UNETR提升了2.1%。此外，Swin*(HybridMIM)实现了95.97%的Dice与0.20的HD95，仅次于UNet(HybridMIM)。
\textbf{Spleen segmentation.} The spleen segmentation results are listed on the right side of Table~\ref{tab:msd_segmentation}.
The HybridMIM based on both UNet and SwinUNETR architectures presented improved performance, both on Dice and HD95. 
HybridMIM(UNet) obtains Dice and HD95 with 96.05 and 0.20, respectively, improving the Dice score by 0.50\% compared to TransVW, and by 2.1\% compared to UNETR. 
%%
%In addition, Swin*(HybridMIM) achieves 95.97\% Dice and 0.20 HD95, second only to UNet (HybridMIM).
% 值得注意的是，SwinUNETR方法的Dice得分为94.61，而我们提出的通用预训练模型Swin* (HybridMIM)方法则达到了95.97的Dice得分，实现了1.36%的提升。通过我们提出的Hybrid的多层次自监督学习方式首先学习丰富的3D脾脏数据的空间解剖学特征，然后通过迁移学习在下游分割任务中训练，可以明显的提升原模型的效果。
Among the fully supervised methods, SwinUNETR gets the best Dice score of 94.61, and HD 0.25.
Our generic pre-trained model HybridMIM(Swin) further improves SwinUNETR to achieve a Dice score of 95.97, realizing an increase of 1.36\%.
%%
%The original model can significantly improve by learning the spatial anatomical features of the rich 3D spleen data through our proposed Hybrid's multi-level self-supervised learning approach and then training it in the downstream segmentation task through transfer learning.

\begin{figure*}[tbp] %H为当前位置，!htb为忽略美学标准，htbp为浮动图形
\vspace{-4mm}
\centering %图片居中
\includegraphics[width=\textwidth]{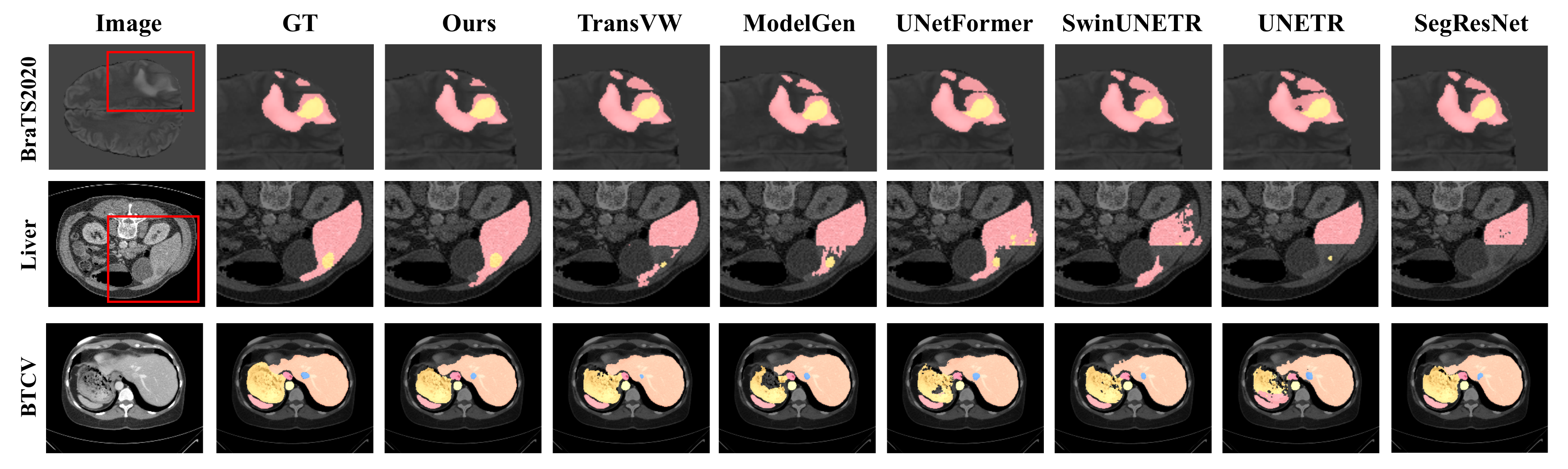} %插入图片，[]中设置图片大小，{}中是图片文件名
% Ours为Swin*(HybridMIM)方法，三行视觉比较结果分别为BraTS2020，Liver和BTCV。我们提出的方法更够更好的分割细微的病灶(第一行)，并且分割的完整度更高(第二行，第三行)。
\vspace{-3mm}
\caption{Qualitative visualizations of the proposed HybridMIM and baseline methods. "Ours" is the HybridMIM(Swin) method. The three rows of visual comparison results are from BraTS2020, Liver, and BTCV datasets. Our proposed method is better for segmenting tiny lesions (first row) and has higher segmentation integrity (second row, third row).} %最终文档中希望显示的图片标题
\label{fig:visual} %用于文内引用的标签
\end{figure*}

% 基于BraTS2020数据的脑胶质瘤的分割结果被展示在表4中。我们使用Dice来评测不同方法的性能。其中WT，TC，ET分别代表了全部肿瘤，肿瘤核心，增强肿瘤，Avg代表3个分割目标的Dice均值。
\textbf{Brain tumor segmentation.} The segmentation results of gliomas for BraTS2020 dataset are summarized in Table \ref{tab:brats_segmentation}. 
%We use Dice to evaluate the performance of different methods. 
WT, TC, ET represent whole tumor region, tumor core, and enhanced tumor region, respectively, and Avg is the Dice mean of the three segmentation targets.
% 我们提出的Swin（MP-SSL）方法实现了一个state-of-the-art的分割结果并且在WT，TC，ET三个分割目标中均达到了最优，分别为91.48%，86.88%，80.81%。相比于没有加入预训练的SwinUNETR方法，Swin（MP-SSL）在三个分割目标中均有较大幅度的提升，分别提升了1.4%，1.69%，0.8%，且三个分割目标的平均Dice得分比第二名TransVW方法提升了0.59%。
Our task-specific pre-trained model HybridMIM*(Swin) reports the best in WT, ET, and Avg with 91.48\%, 80.81\%, and 86.39\% respectively.
% 对比没有预训练的SwinUNETR方法，Swin(HybridMIM)与Swin* (HybridMIM)在三个分割目标上均有较大的提升，相比SwinUNETR，平均的Dice分别提升了1.3%, 1.24%。
%Compared with the SwinUNETR method without pre-training, Swin(HybridMIM) and Swin* (HybridMIM) show a considerable improvement in all three segmentation objectives, with an average Dice improvement of 1.3\%, 1.24\%, respectively, compared to SwinUNETR.
% 此外，UNet方法经过预训练后，也有了非常明显的提升，像表中最后一行展示的那样，UNet* (HybridMIM)方法在三个分割目标7分别实现了90.41%， 86.49%， 80.61%的Dice得分，相比于同样为UNet架构的ModelGen，三个分割指标的平均Dice提升了0.12%。以上的结果充分证明了MP-SSL方法良好的迁移学习和模型泛化能力。
%%In addition, the UNet method shows a significant improvement after pre-training, as shown in the last row of the table. 
As for UNet as the underlying architecture, the generic pre-trained model HybridMIM(UNet) achieves Dice scores of 90.41\%, 86.49\%, and 80.61\% for the three segmentation targets, respectively. Compared with ModelGen which is also built on UNet, we has the average Dice improved by 0.12\%. 
%%
%The above results fully demonstrate the good transfer learning and model generalization ability of the HybridMIM method.
It is also noted that on BraTS2020 dataset, the task-specific pre-trained mode gets better performance than the generic pre-trained mode. 

\begin{table}[t]
    \centering
    % BraTS2020数据集包含四个模态，三个分割目标。我们选择UNet和SwinTransformer作为backbone，分别于有监督学习方法跟自监督学习方法对比，结果展示了UniLearn对不同架构的有效性。
    \caption{Quantitative comparison on BraTS 2020 dataset, which contains four modalities and three segmentation targets. }
    % \vspace{-3mm}
    \label{tab:brats_segmentation}
    \renewcommand\arraystretch{1.3}
    \setlength\tabcolsep{10pt}%调列距
    \resizebox{0.48\textwidth}{!}{
    \begin{tabular}{c | c c c c}
    \hline
    Methods & WT & TC & ET & Avg\\
    \hline
    SegresNet & 90.04 & 85.08 & 78.81 & 84.64 \\
    
    UNETR & 89.92 & 84.79 & 79.51 & 84.74\\
    SwinUNETR & 90.08 & 85.19 & 80.01 & 85.09\\
    \hline
    ModelGen & 90.60 & 86.59 & 79.95 & 85.71\\
    TransVW & 90.96 & 86.26 & 80.20 & 85.80 \\
    UNetFormer* & 90.93 & 86.17 & 79.97 & 85.69\\
    UNetFormer & 90.71 & 86.22 & 80.19 & 85.71\\
    \hline
    HybridMIM*(Swin) & \textbf{91.48} & {86.88} & \textbf{80.81} & \textbf{86.39} \\
    HybridMIM*(UNet) & 90.62 & 86.28 & 80.17 & 85.69\\
    \hline
    HybridMIM(Swin) & 90.95 & \textbf{87.34} & 80.71 & 86.33\\
    HybridMIM(UNet) & 90.41 & 86.49 & 80.61 & 85.83 \\
    \hline
    \end{tabular}
    }
    \vspace{-2mm}
\end{table}

\begin{figure}[htbp] %H为当前位置，!htb为忽略美学标准，htbp为浮动图形
\centering %图片居中
\vspace{-2mm}
\includegraphics[width=0.8\columnwidth]{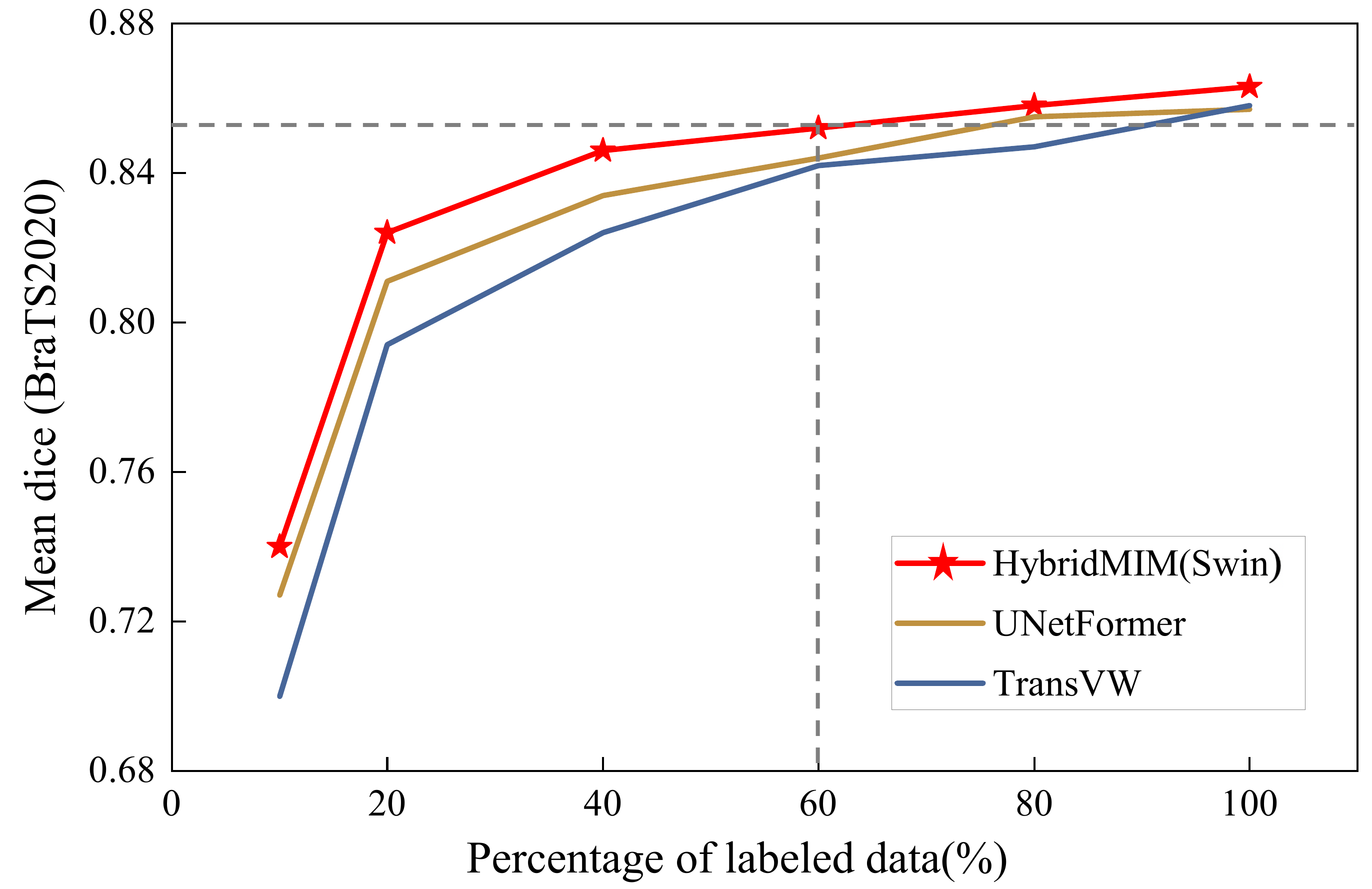} %插入图片，[]中设置图片大小，{}中是图片文件名
% 不同有标签数据规模对迁移学习结果的影响。我们分别选择了BraTS2020数据集中训练数据的10%，20%，40%，60%，80%，100%，验证在不同自监督学习方法的迁移学习能力。
\caption{Effect of different labeled data sizes on migration learning results. We selected 10\%, 20\%, 40\%, 60\%, 80\%, and 100\% of the training data in the BraTS2020 dataset to verify the transfer learning ability in different self-supervised learning methods.} %最终文档中希望显示的图片标题
\label{fig:data_proportion}
\vspace{-2mm}
%用于文内引用的标签
\end{figure}

\vspace{-2mm}
\subsection{Qualitative Comparison to Previous Methods}

% 为了更加直观的对比不同方法的分割结果，我们选择Swin*(HybirdMIM)和其他六个性能较好的对比方法在BraTS2020，Liver和BTCV数据集上进行视觉比较。
To compare the segmentation results of different methods more intuitively, we choose HybridMIM(Swin) and four comparative methods with better performance on the BraTS2020, Liver, and BTCV datasets for visual comparison.
% 像Fig. 6. 所展示的，Swin*(HybridMIM)能够提升病灶识别的准确度和完整度，并且针对细微的病灶依然可以高效的识别出来。模型经过HybridMIM方法预训练后，对局部区域的感知能力更强。
As shown in Figure~\ref{fig:visual}, HybridMIM(Swin) can improve the accuracy and completeness of lesion identification,  and still perceive subtle lesions. 
%The model is pre-trained by the HybridMIM method and better perceives localized regions.
%在Fig. 6. 的第一行，可以明显看出我们的方法相比于其他对比方法可以更加精准的分割微小的病灶。在Liver数据集中（Fig. 6.第二行），Swin*(HybridMIM)分割的完整性更高，没有出现像其他对比方法中的分割区域不连续的情况。同时，在BTCV数据集中的可视化结果中，我们的方法的分割结果包含的空洞更少，与其他对比方法相比，有较高的完整度。 
To be specific, for brain tumor in BraTS2020 (the first row of Figure~\ref{fig:visual}), our method segments the whole tumor with more accurate boundary, while the comparative methods all enlarge the tumor region. 
In the liver segmentation task (the second row), we can clear see that the comparative methods generate obvious discontinuity in the segmented areas. Especially UNETR and SegResNet fail to detect the lower part of the liver, while the detected liver region from our method exhibits a clearly higher integrity. 
For the BTCV dataset, TransVW, UNetFormer, SiwnUNETR generates small holes in stomach; ModelGen even is subjected to a much large missing detected part. In contrast, our segmentation result is more close to the ground truth.

\vspace{-2mm}
\subsection{Reduce Manual Labeling Efforts}
% 为了验证随着有标签数据比例逐渐降低，HybridMIM方法相比于其他自监督学习方法依然能保持良好的迁移学习能力，我们选择UNetFormer与TransVW作为对比方法，BraTS2020作为下游分割任务数据集，采用10%，20%，40%，60%，80%，100%的数据比例进行对比实验。
To evaluate the transfer learning ability with annotation scarcity challenge in medical imaging, we conduct the experiment of finetuning using a subset of BraTS2020 data.  
Figure~\ref{fig:data_proportion} demonstrates the comparison results between HybridMIM(Swin), TransVW and UNetFormer. 
%%
%In order to verify that as the proportion of labeled data gradually decreases, the HybridMIM method still maintains good transfer learning ability. We choose UNetFormer and TransVW as the comparison methods and BraTS2020 as the downstream segmentation task dataset and use 10\%, 20\%, 40\%, 60\%, 80\%, and 100\% data proportions for comparison experiments.
% Fig. 4. 展示了减少有标签数据比例的实验结果。实验结果表明，当有标签数据比例降低至60%时，UNetFormer与TransVW方法在BraTS2020分割数据集上的迁移学习能力明显降低。而通过HybridMIM方法预训练的通用模型SwinUNETR在有标签数据比例为20%时依然能够实现0.825的平均Dice。
%Fig. \ref{data_proportion} shows the experimental results of reducing the proportion of labeled data.
It is clear that the generic pre-trained model HybridMIM(Swin) presents the best performance when using the same portion of labelled data.
On employing 20\% labelled data, HybridMIM(Swin) already achieves an average Dice of 82.55\%, with 1.42\% and 3.17\% higher than UNetFormer and TransVW, respectively.  
The Dice 85.24\% can be achieved by using HybridMIM(Swin) with 60\% labelled data, while UNetFormer requires about 80\% data and TransVW requires nearly 90\% data.
%%
%%On employing 40\% labelled data, HybridMIM(Swin) obtains an average Dice of ??, even higher than UNetFormer and TransVW employing 60\% labelled data. 
 
%The experimental results show that the transfer learning ability of UNetFormer and TransVW methods declined significantly on the BraTS2020 segmented dataset when reducing the proportion of labeled data to 60\%. In contrast, Swin, a generic model pre-trained by the HybridMIM method, still achieves an average Dice of 0.825 when the proportion of labeled data is 20\%.
% 此外，当有标签数据的比例相同时，Swin*(HybridMIM)较其他对比方法均有明显的性能优势。并且Swin*(HybridMIM)需要更少的数据便可以实现其他对比方法需要更多数据才能实现的性能，例如Swin*(HybridMIM)利用60%的有标签数据达到的迁移学习的性能，UNetFormer需要80%的数据，TransVW需要90%的数据。
%In addition, the HybridMIM(Swin) has a significant performance advantage over other comparison methods when the proportion of labeled data is the same. For example, the HybridMIM(Swin) achieves transfer learning performance with 60\% of labeled data, while UNetFormer requires 80\% of data and TransVW requires 90\% of data.

\vspace{-2mm}
\subsection{Pre-training Speed Comparison}
\begin{figure}[htbp] %H为当前位置，!htb为忽略美学标准，htbp为浮动图形
\centering %图片居中
\vspace{-2mm}
\includegraphics[width=0.8\columnwidth]{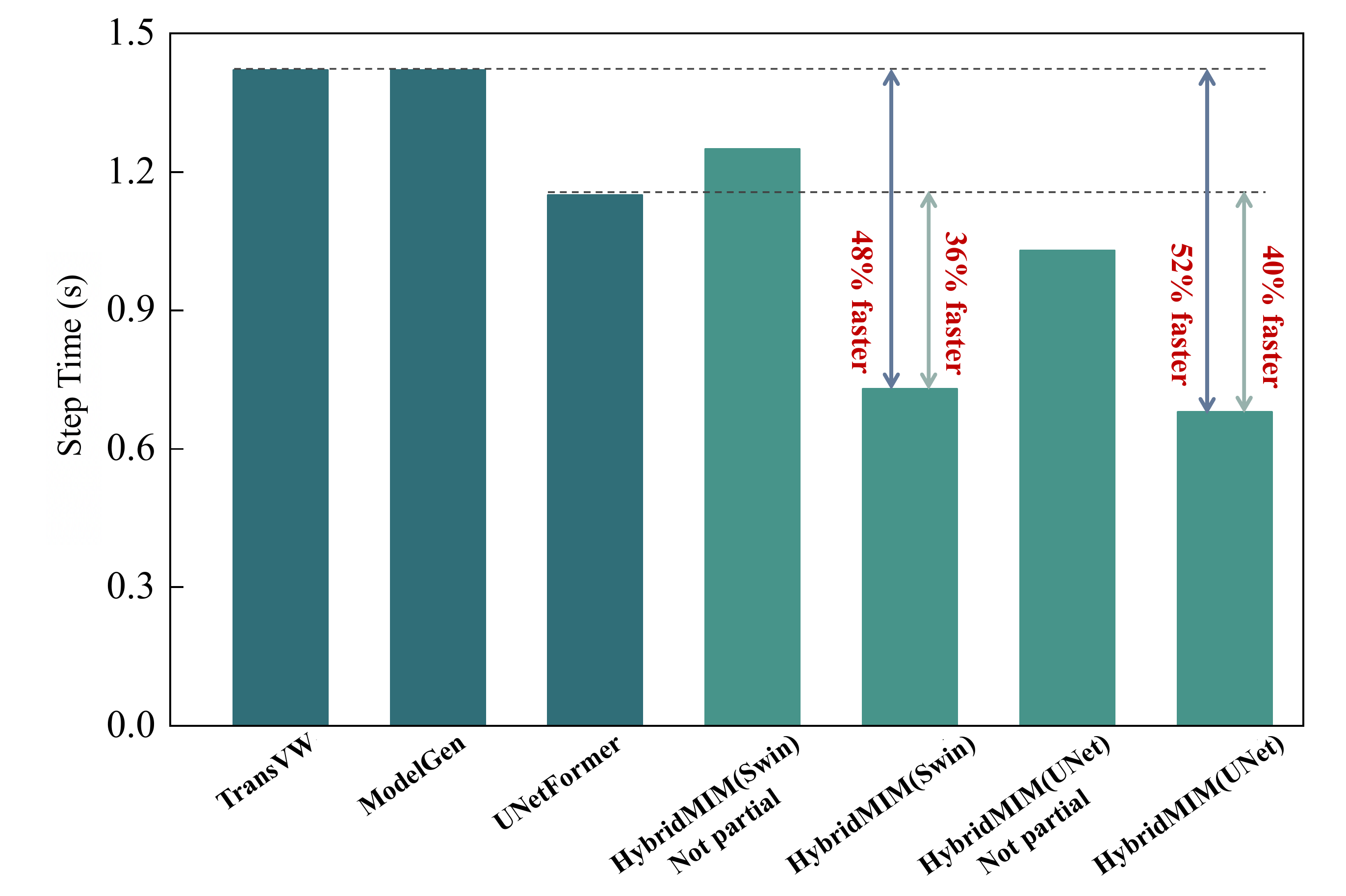} %插入图片，[]中设置图片大小，{}中是图片文件名
% 不同自监督学习方法预训练时间消耗对比。横坐标为不同自监督学习方法和不同重建大小的HybridMIM方法，128是全局重建大小，96是我们提出的局部重建方式。纵坐标表示预训练时每步的时间消耗。
\caption{Comparison of pre-training time consumption for different SSL methods. 
%The horizontal coordinates are different self-supervised learning methods. 
``Not partial'' denotes that the partial region prediction scheme is not used.
%, which spend more time in pre-training. The vertical coordinate indicates the time consumption of each step during pre-training.
} %最终文档中希望显示的图片标题
\label{fig:pretraining_time} %用于文内引用的标签
\vspace{-2mm}
\end{figure}

% 在自监督学习的过程中，由于无标签数据的数据量通常较大，因此训练速度是一个影响自监督学习方法的非常重要的因素。MP-SSL通过灵活的选择局部的一级区域重建来提升预训练速度。我们与其他的自监督学习方法进行对比，像图3(d)中展示的那样，我们分别列举了基于UNet与SwinTransformer架构的MP-SSL方法与其他自监督方法的时间消耗。
In self-supervised learning, the training speed is a notable factor to consider, because the unlabeled data scale is usually large especially in the generic training mode. 
Figure~\ref{fig:pretraining_time} demonstrates the time consumption of those self-supervised methods in the pre-training stage on BraTS2020 dataset.
%%
%The HybridMIM enhances the pre-training speed by flexibly selecting local first-level region reconstruction. We compare with other self-supervised learning methods, as shown in Fig. \ref{pretraining_time}, and we enumerate the time consumption of the HybridMIM method based on UNet and SWinUNETR architectures, respectively, with other self-supervised methods.
% 值得注意的是，为了更加公平的进行对比，我们对比了每个自监督学习方法运行一步的平均时间消耗。其中一步内包含了前向传播，反向传播，更新参数，而不包含数据读取，数据预处理等时间消耗不确定的操作。
For a fair comparison, we count the average time of running one step for each method, which contains forward prediction, backward propagation, and updating network parameters, but does not include data reading and preprocessing operations.

% 因此，由图3(d)可以看出，TransVW与ModelGenesis方法时间消耗最多。Swin（HybridMIM）当使用（128，128，128）作为重构区域时，由于其包含更多的损失函数，因此时间消耗高于类似架构的UNetFormer方法。但是随着我们将需要重构的局部区域降低为（96，96，96），预训练时间大幅度降低，相比于TransVW与ModelGen方法，预训练速度提升48%，相比于UNetFormer方法，预训练速度提升36%。
As Figure \ref{fig:pretraining_time} shows, the TransVW and ModelGenesis methods with the same underlying architecture have the highest time consumption, both of which are 1.42s per step. 
HybridMIM(Swin), when predicting all the masked sub-volumes (denoted as ``Not partial''; see the fourth bar), has a higher time consumption than the UNetFormer method. 
It is because that although they have the similar underlying architecture, HybridMIM(Swin) involves  more loss functions. 
On the other hand, when we apply the partial region prediction, the pre-training time of HybridMIM(Swin) decreases dramatically, in which the speedup is 48\% with respect to TransVW and ModelGen, and 36\% against UNetFormer.

% 类似的，当使用UNet(HybridMIM)方法时，此时虽然由于所使用的UNet本身的结构特殊性，有更低时间消耗，但通过选择局部区域重建，训练速度依然有显著的提升。像表3d中展示的那样，当使用(128,128,128)大小作为重构尺寸时，每步时间消耗为1.03s，而当使用（96，96，96）大小时，每步时间消耗降低0.35s，相比TransVW和ModelGen方法，预训练速度快52%，相比UNetFormer方法，预训练速度加快40%。
When using the HybridMIM(UNet) method, there is a lower time consumption due to the structural simplicity of the UNet (see the rightmost two bars). 
The partial region prediction enables it to get a significant improvement in the pre-training speed, with the time consumption per step reduced by 0.35s.
HybridMIM(UNet) achieves a pre-training  speed of 0.68s, 52\% faster than the TransVW and ModelsGenesis methods, and 40\% faster than the UNetFormer method.
%%
% It is worthy noting that the pre-training speed is close to the training speed in the finetuning, despite that the later has fewer losses to compute.
% %%
% Therefore, our method can also have faster time performance in the finetuning stage.

\vspace{-2mm}
\subsection{Ablation Study}
\subsubsection{Selection of the optimal architecture settings}
%\vspace{-4m}
\begin{figure}[htbp] %H为当前位置，!htb为忽略美学标准，htbp为浮动图形
\vspace{-4mm}
\centering %图片居中
\includegraphics[width=0.48\textwidth]{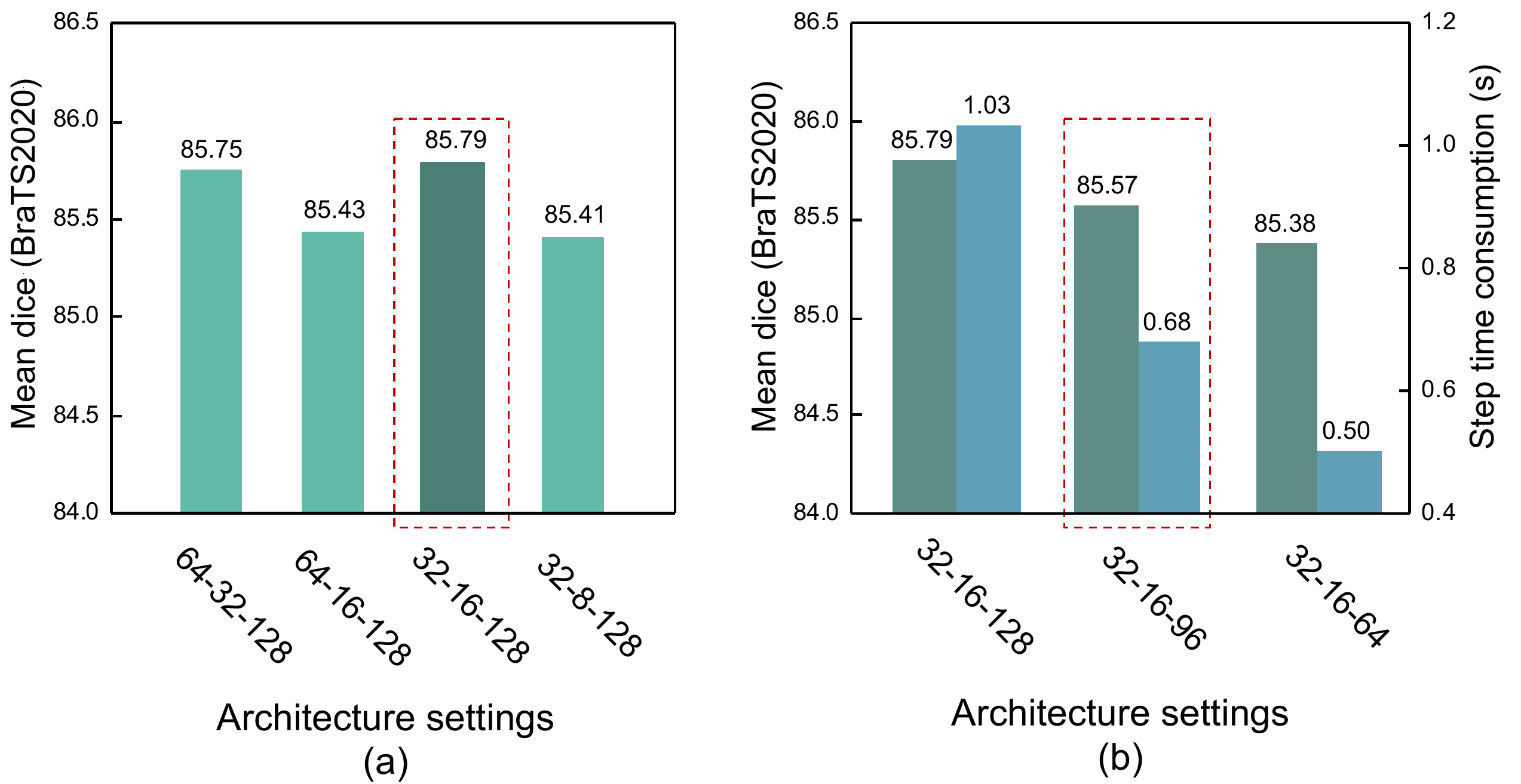} %插入图片，[]中设置图片大小，{}中是图片文件名
% 不同架构参数对迁移学习性能与预训练时间的影响。(a)中横坐标中a-b-c分别代表一级区域大小，二级区域大小，重建区域大小。纵坐标表示在BraTS2020数据集迁移学习能力（三个分割目标的Dice平均值）。(b)中右侧纵坐标表示预训练时每个step消耗的时间。我们首先通过(a)确定最优的一级区域与二级区域，32-16-128迁移学习效果最好。之后，我们通过(b)改变重建区域的大小，兼顾性能与时间选择最优的架构参数设置。
\caption{Effect of different architecture parameters on transfer learning performance and pre-training time. The a-b-c in the horizontal coordinates in (a) represent the first-level, second-level, and reconstructed region sizes, respectively. The vertical coordinates represent the transfer learning capability in the BraTS2020 dataset (average Dice for the three segmentation targets). The right vertical coordinate in (b) indicates the time consumed per step during pre-training. The two red dashed boxes indicate the optimal architectural parameters we choose in (a) and (b), respectively. } %最终文档中希望显示的图片标题
%% 两个红色虚线框分别表示了我们在(a)和(b)中选择的最优架构参数。
%% We determine the optimal first-level and second-level regions by (a), and 32-16-128 migration learning works best. After that, we change the size of the reconstructed region by (b) choosing the optimal architecture parameter settings considering the performance and time.
\label{fig:pretraining_setting}
\vspace{-2mm}
%用于文内引用的标签
\end{figure}
%\label{pretraining_settings}

% 为了选择一个更好的架构参数，我们进行了多组对照实验。我们选择UNet架构预训练多组通用模型，see Fi. 3. 横坐标架构设置a-b-c中，a表示一级区域的大小，b表示二级区域的大小，c表示重建大小。纵坐标为通用模型在BraTS2020数据集中finetuning的Dice指标。

%%
In order to choose an optimal architecture setting, we conduct a multigroup control experiment. 
We choose the UNet architecture to pre-train the possible settings (see Figure~\ref{fig:pretraining_setting}), where the three numbers under each bar represent the first-level sub-volume size, the second-level patch size, and the region size for partial region prediction.  
The left vertical coordinates are the Dice metrics of finetuning the generic pre-trained model on the BraTS2020 dataset.
% Fig. 3. (b)中右侧纵坐标为每个step的时间消耗。
The right vertical coordinate in Figure~\ref{fig:pretraining_setting} (b) is the time consumption of each pre-training step.
% 像Fig. 3.(a)中所展示的那样，我们固定预训练的重构大小为128，选取了64-32，64-16，32-16，32-8四组参数预训练通用模型，之后在BraTS2020分割任务中进行finetuning，结果显示，32-16-128的参数设置表现最好，实现了最好的Dice。

As Figure~\ref{fig:pretraining_setting} (a) shows, we first fix the region size for partial region prediction to be 128, select four sets of parameters (64-32, 64-16, 32-16, and 32-8) for sub-volume and patch sizes.
%, and later perform finetuning in the BraTS2020 segmentation task. 
The results show that the parameter setting of 32-16-128 performs the best and achieves the best Dice of 85.79\%.

% 之后，我们选择32-16参数设置，逐步减小重构大小，see Fig. 3. (b)，实验结果展示，重建大小由128降低到96时，每个step的时间由1.03s降低至0.68s。下游分割任务的Dice指标由0.875降低至0.860。当重建大小继续降低至64时，每个step的时间为0.50s，Dice指标为85.38。为了实现更快的预训练速度并使性能影响降低，我们选择32-16-96作为我们的架构参数设置。^^
Afterwards, we fix the optimal sub-volume and patch sizes (32-16), and gradually decrease the reconstruction region size; see Figure~\ref{fig:pretraining_setting} (b). 
We can see that with a smaller reconstruction region size, the Dice score decreases a little bit, while the time performance reduces greatly. 
For instance, when reducing the reconstruction size from 128 to 96, the Dice score for the downstream segmentation task decreases from 85.79\% to 85.57\%, and the time per step decreases from 1.03s to 0.68s. 
%when the reconstruction size decreases to 64, the time per step is 0.50s, and the Dice metric is 85.38. The Dice metric is 85.38 for 0.50s. 
Considering the trade-off between the segmentation accuracy and pre-training speed, we choose 32-16-96 as our architecture parameters for the case that the input sample has a size of $128\times128\times128$ (BraTS2020 dataset).
Taking this experiments as guidance, we use an architectural parameter setting of 32-16-64 for the case that the input sample has a size of $96\times96\times96$ (BTCV, MSD Liver and MSD Spleen).

% 我们分别使用了UNet与SwinTransformer作为backbone，在BraTS2020数据集上通过消融实验充分的验证了我们提出的每个模块的有效性。实验结果被展示在表5中。Loss单元格包含五个不同的损失函数，分别为LR(local reconstruction), Num(number), Loc(location), Consis(consistency), CL(contrastive learning)，其中LR代表了像素层次的3D医学图像表征的学习，Num，Loc，Consis代表了区域层次的表征学习，而CL代表了样本层次的学习。我们验证了MP-SSL在不同层次上的自监督学习对下游分割任务的性能提升。
\subsubsection{Efficiency of Self-Supervised Objectives}

We comprehensively validate the effectiveness of our modules through ablation experiments on the BraTS2020 dataset. 
The experimental results using the generic pre-training mode are presented in Table~\ref{tab:ablation}. 
We have five loss functions, namely $\mathcal{L}_{\mathrm{PR}}$ (partial region prediction), $\mathcal{L}_{\mathrm{Num}}$ (number prediction), $\mathcal{L}_{\mathrm{Loc}}$ (location prediction), $\mathcal{L}_{\mathrm{Con}}$ (consistency between number and location prediction), and $\mathcal{L}_{\mathrm{CL}}$ (contrastive learning).
$\mathcal{L}_{\mathrm{PR}}$ facilitates the learning of 3D medical image latent representations at the pixel level; the combination of $\mathcal{L}_{\mathrm{Num}}$, $\mathcal{L}_{\mathrm{Loc}}$, and $\mathcal{L}_{\mathrm{Con}}$ facilitates the learning at the region level; and $\mathcal{L}_{\mathrm{CL}}$ facilitates the learning at the sample level. 
%We validate the performance improvement of the HybridMIM method at different levels of self-supervised learning for downstream segmentation tasks.
% Segmentation Target表示BraTS2020数据集不同的分割目标，Avg代表三个分割目标的平均指标。
%Segmentation Target represents the different segmentation targets of the BraTS2020 dataset, and Avg represents the average metric of the three segmentation targets.
% 表格中每个backbone的第一行结果为基线，不进行预训练，而是直接在下游分割任务上进行训练。之后，我们在预训练过程中逐渐添加不同的损失函数，来验证我们提出的不同模块对不同网络架构的性能提升能力。
We make comparison to the baseline with supervised training from scratch on the BraTS2020 dataset (see the first row for each backbone). 
%After that, we gradually add different loss functions during the pre-training process to verify the performance improvement capability of our proposed different modules for different network architectures.

\begin{table}[th]
    \centering
    \vspace{-3mm}
    % 在BraTS2020数据集上进行消融实验。我们选择UNet与SwinTransformer作为backbone，逐个添加我们提出的不同层次的损失函数。其中LR为局部重建损失，Num为数量分布预测损失，Loc为位置分布预测损失，Consis为一致性损失，CL为对比学习损失。下游任务的分割结果展示了我们提出的每个损失函数对于不同架构的有效性。
    \caption{Ablation experiments are performed on the BraTS2020 dataset. $\mathcal{L}_{\mathrm{LR}}$: the local reconstruction loss, $\mathcal{L}_{\mathrm{Num}}$: the number distribution prediction loss, $\mathcal{L}_{\mathrm{Loc}}$: the location distribution prediction loss, $\mathcal{L}_{\mathrm{Con}}$: the consistency loss, $\mathcal{L}_{\mathrm{CL}}$: the contrastive learning loss. }
    %The segmentation results of the downstream task demonstrate the effectiveness of each of our proposed loss functions for different architectures.
    % \vspace{-3mm}
    \label{tab:ablation}
    \renewcommand\arraystretch{1.2}
    \setlength\tabcolsep{5pt}%调列距
    \resizebox{\columnwidth}{!}{
    \begin{tabular}{c| l | c c c c c c}

    \hline
    \multirow{2}*{\makecell{Backbone}} & \multirow{2}*{Loss} & \multicolumn{4}{c}{Segmentation Target} & \\
    % \cline{3-7} \cline{10-13}
     & &  WT & TC & ET & Avg & \\
    \hline
    %% LR & Num & Loc & Consis & CL
    \multirow{6}{*}{UNet} & Supervised learning & 89.75 & 84.65 & 78.83 & 84.41 &\\
     & $\mathcal{L}_{\mathrm{PR}}$ & 90.19 & 85.50 & 79.48 & 85.06 & \\
     & $\mathcal{L}_{\mathrm{PR}} + \mathcal{L}_{\mathrm{Num}}$ & 90.05 & 85.48 & 79.97 & 85.17 & \\
     & $\mathcal{L}_{\mathrm{PR}} + \mathcal{L}_{\mathrm{Num}} + \mathcal{L}_{\mathrm{Loc}}$ & 90.15 & 85.65 & 80.10 & 85.30 & \\
     & $\mathcal{L}_{\mathrm{PR}} + \mathcal{L}_{\mathrm{Num}} + \mathcal{L}_{\mathrm{Loc}} + \mathcal{L}_{\mathrm{Con}}$ & 90.30 & 85.36 & \textbf{80.56} & 85.40 & \\
     & $\mathcal{L}_{\mathrm{PR}} + \mathcal{L}_{\mathrm{Num}} + \mathcal{L}_{\mathrm{Loc}} + \mathcal{L}_{\mathrm{Con}} + \mathcal{L}_{\mathrm{CL}}$ & \textbf{90.62} & \textbf{86.28} & {80.17} & \textbf{85.69} & \\
     \hline
     
     \multirow{6}{*}{Swin} & Supervised learning & 90.08 & 85.19 & 80.01 & 85.09 &\\
     & $\mathcal{L}_{\mathrm{PR}}$ & 90.95 & 86.17 & 80.22 & 85.78 & \\
     & $\mathcal{L}_{\mathrm{PR}} + \mathcal{L}_{\mathrm{Num}}$ & 90.93 & 86.94 & 80.48 & 86.12 & \\
     & $\mathcal{L}_{\mathrm{PR}} + \mathcal{L}_{\mathrm{Num}} + \mathcal{L}_{\mathrm{Loc}}$ & 91.18 & 86.33 & \textbf{81.10} & 86.20 & \\
     & $\mathcal{L}_{\mathrm{PR}} + \mathcal{L}_{\mathrm{Num}} + \mathcal{L}_{\mathrm{Loc}}  + \mathcal{L}_{\mathrm{Con}}$ & 90.98 & \textbf{87.06} & 80.71 & 86.24 & \\
     & $\mathcal{L}_{\mathrm{PR}} + \mathcal{L}_{\mathrm{Num}} + \mathcal{L}_{\mathrm{Loc}}  + \mathcal{L}_{\mathrm{Con}} + \mathcal{L}_{\mathrm{CL}}$ & \textbf{91.48} & {86.88} & {80.81} & \textbf{86.39} & \\
    %  \hline

    % \multirow{4}{*}{Swin} & & &  & & & & & & 90.08 & 85.19 & 80.01 & 85.09 & \\
    %  & & \checkmark & & & &  & & & 90.95 & 86.17 & 80.22 & 85.78 & \\
    %  & & \checkmark & \checkmark & & &  & & & 90.93 & 86.94 & 80.48 & 86.12 & \\
    %  & & \checkmark & \checkmark & \checkmark & &  & & & 91.18 & 86.33 & 81.10 & 86.20 & \\
    %  & & \checkmark & \checkmark & \checkmark & \checkmark &  & & & 90.98 & 87.06 & 80.71 & 86.24 & \\
    %  & & \checkmark & \checkmark &\checkmark & \checkmark & \checkmark & & & {91.48} & {86.88} & {80.81} & {86.39} & \\
    \hline
    \end{tabular}
    }
    \vspace{-2mm}
\end{table}

% 从表5中可以清晰的看出，当使用UNet架构在BraTS2020数据集上从零开始训练时，三个分割目标的Dice得分分别为89.75%， 84.65%， 78.83%， 平均值为84.41%。
\textbf{UNet architecture.} The baseline that is trained from scratch reports the Dice scores 89.75\%, 84.65\%, and 78.83\%, for the three segmentation targets respectively, with an average number of 84.41\%. 
% 此时加入第一个自监督学习损失LR(local reconstruction)，该损失从像素层次来重建原图像被掩蔽区域的分布。在下游分割任务上加载由LR损失预训练得到的模型权重，使得每项分割目标均有不同程度的提升，平均值达到85.06%，较从零开始训练提升了0.65%。
At this point, we add the first self-supervised learning loss $\mathcal{L}_{\mathrm{PR}}$, which reconstructs the masked regions of the original image at the pixel level. The model weights fine-tuned onto the downstream segmentation task, result in a Dice average of 85.06\%, with an improvement of 0.65\% over the baseline.
% 之后，添加区域层次的自监督损失Num(number),Loc(location),Consis(consistency)，提升模型表征空间区域分布的能力，分割目标的均值由85.06%提升至85.40%。
The addition of region-perception losses, i.e. $\mathcal{L}_{\mathrm{Num}}$, $\mathcal{L}_{\mathrm{Loc}}$, $\mathcal{L}_{\mathrm{Con}}$, improves the model's ability to characterize the distribution of spatial regions, and the mean Dice value is increased from 85.06\% to 85.40\%, getting an improvement of 0.34\%.
% 最后，添加样本层次的自监督损失CL(contrastive learning)，提升模型对于不同样本表征的区分能力。通过CL损失，在下游分割任务中，三个分割目标的Dice得分均值达到了85.69%，并且在WT与TC上的Dice得分也达到了最高，分别为90.62%和86.28%。
Finally, we add the sample-level self-supervised loss $\mathcal{L}_{\mathrm{CL}}$ to enhance the model's ability to distinguish between different sample representations. With $\mathcal{L}_{\mathrm{CL}}$, the mean Dice score reaches 85.69\% in the downstream segmentation task, and the highest Dice scores of 90.62\% and 86.28\% on WT and TC, respectively. 
In the end, the average Dice score with pre-training was 1.29\% higher than that without pre-training.

% 类似的，MP-SSL方法对于SwinTransformer架构也有较大程度的提升。三个分割目标的平均Dice得分由没有预训练时候的85.09%最终提升到了86.39%，在BraTS2020数据集上实现了SOTA的分割结果。
\textbf{SwinUNETR architecture.} Similarly, the HybridMIM method also achieves obvious improvements for the SwinUNETR architecture. The average Dice score of the three segmentation targets was finally improved from 85.09\% without pre-training to 86.39\%, achieving SOTA segmentation results on the BraTS2020 dataset. 

% \textbf{Analysis of self-supervised loss enhancement effects.} 对于UNet跟SwinTransformer架构，从表中可以看出，LR损失发挥了比较大的作用。UNet架构加入LR损失后，三个分割指标的平均Dice得分提升了0.65%，而SwinTransformer架构加入LR损失后，三个指标的平均Dice得分提升了0.69%。
\textbf{Analysis of self-supervised loss enhancement effects.} 
For the UNet and SwinTransformer architectures, Table~\ref{tab:ablation} shows that the $\mathcal{L}_{\mathrm{PR}}$ plays a larger role. The average Dice score of the three segmentation targets increases by 0.65\% with the aid of $\mathcal{L}_{\mathrm{PR}}$ upon the UNet architecture, while the average Dice score of the three metrics increased by 0.69\% upon the SwinTransformer architecture.
% 此外Consis损失由于具有保持预测的数量与位置信息一致的作用，提升自监督学习的可解释性，因此其对于下游分割任务的提升较小。对于UNet架构，平均Dice得分提升了0.1%，而对于SwinTransformer结构，平均Dice提升了0.04%。
%%
The region perception losses ($\mathcal{L}_{\mathrm{Num}}$, $\mathcal{L}_{\mathrm{Loc}}$, $\mathcal{L}_{\mathrm{Con}}$ together) are the second important. 
Also note that although the $\mathcal{L}_{\mathrm{Con}}$ has a relatively small improvement for the downstream segmentation task, it has a role in keeping the predicted quantity consistent with the location information, improving the interpretability of the self-supervised learning. 
%For the UNet architecture, the average Dice score increased by 0.1\%, while for the SwinTransformer structure, the average Dice increased by 0.04\%.

\section{Conclusion}

% 我们提出了一个通用的预训练框架(UniLearn)针对于3D医学图像分割，既可以支持卷积神经网络结构又可以支持Transformer结构。UniLearn的核心是一个多层次的自监督学习策略，分别从像素层次，区域层次，样本层次学习3D医学图像的解剖学特征。我们利用多个器官的分割数据集，选择了有监督学习和自监督学习方法进行对比实验。实验结果表明了经过UniLearn进行预训练后，原架构可以得到较大的性能提升，并实现state-of-the-art的分割结果。
In this paper, we propose a Hybrid Masked Image Modeling framework for 3D medical image segmentation pre-training, which can support CNN and Transformer structures. 
The core of HybridMIM is a multi-level SSL strategy that learns anatomical features of 3D medical images from the pixel-level, region-level, and sample-level, respectively. 
% 相比于其他自监督学习方法，HybridMIM通过掩蔽式图像建模的方式更高效的在像素维度学习高维医学图像数据的解剖特征。同时也通过分层的掩蔽方式，在区域维度学习医学图像数据的空间分布。最后，HybridMIM利用对比学习在样本维度增强模型对于不同数据的区分能力。此外，局部重建等优化方式的加入，也提升了预训练效率，减小预训练模型的成本。
Compared with other SSL methods, HybridMIM learns spatial anatomical features of medical image data more effectively through a hierarchical masking approach, which supports pixel-level partial region prediction and region-level patch-masking perception.
%%
%It also learns the spatial distribution of image data more efficiently at the pixel-level by partial region prediction. 
%%
In addition, HybridMIM uses dropout-based contrastive learning at the sample level to enhance the model's ability to discriminate between different data. 
%In addition, adding optimization methods such as local reconstruction also improves the pre-training efficiency and reduces the cost of pre-training models.

We conduct a thorough comparison with both state-of-the-art SSL methods and supervised learning using segmentation datasets of multiple organs. The experimental results show that after pre-training by HybridMIM, the original network can get outperformed segmentation results.
% 紧接着，我们还通过完备的消融实验，在UNet架构与SwinTransformer架构上证明了UniLearn中每个损失函数的有效性。
We also demonstrate the effectiveness of each loss function in HybridMIM on the UNet architecture and SwinUNETR architecture through comprehensive ablation experiments, 
% 此外，预训练通常需要训练大量数据与大量的步数，因此训练效率是十分重要的。我们设计实验对比了UniLearn与其他自监督学习方法的预训练时间消耗，证明了UniLearn的预训练速度优势。
%In addition, pre-training usually requires training a large amount of data with a large number of steps, so training efficiency is very important. 
and demonstrate the pre-training speed advantage of the proposed HybridMIM.
In the future, we will apply HybridMIM pre-training on more architectures to support other downstream tasks.
%learn the semantic representation of medical images. %Meanwhile, we will also collect more datasets on the pre-training phrase to achieve a higher segmentation accuracy on the downstream tasks.

% \textbf{Acknowledgments.}
% This work was supported by the grant from Tianjin Natural Science Foundation (Grant No. 20JCYBJC00960) and HKU Seed
% Fund for Basic Research (Project No. 202111159073).

\bibliographystyle{IEEEtran}
\bibliography{references}

\end{document}